\documentclass[journal]{IEEEtran}
\ifCLASSINFOpdf
\else
\fi

\usepackage{blindtext}
\usepackage{xcolor}
\usepackage{hyperref}
\usepackage{graphicx}
\graphicspath{ {./images/} }
\usepackage{diagbox}
\usepackage{amsmath} 
\usepackage{mathtools}
\usepackage{lipsum}
\usepackage[nocompress,space]{cite}
\usepackage{amssymb}
\usepackage{pdfrender}
\usepackage{blindtext}
\usepackage{xcolor}
\usepackage{hyperref}
\usepackage{graphicx}
\graphicspath{ {./images/} }
\usepackage{diagbox}
\usepackage{mathtools}
\usepackage{lipsum}
\usepackage[nocompress,space]{cite}
\usepackage{pdfrender}
\usepackage[
  separate-uncertainty = true,
  multi-part-units = repeat
]{siunitx}
\usepackage{textcomp}
\usepackage{siunitx}
\usepackage{float}
\usepackage{caption}
\usepackage{adjustbox}
\usepackage{rotating}
\usepackage{blindtext}
\usepackage{xcolor}
\usepackage{hyperref}
\usepackage{graphicx}
\graphicspath{ {./images/} }
\usepackage{diagbox}
\usepackage{amsmath}
\usepackage{mathtools}
\usepackage{lipsum}
\usepackage[nocompress,space]{cite}
\usepackage{pdfrender}
\usepackage{soul,color}
\usepackage{orcidlink}

\usepackage[ruled,vlined]{algorithm2e}

\SetCommentSty{mycommfont}

\IEEEoverridecommandlockouts

\usepackage{xcolor,soul,framed} 

\colorlet{shadecolor}{yellow}
\usepackage{array}
\usepackage{mdwmath}
\usepackage{mdwtab}
\usepackage{eqparbox}
\usepackage{url}

\usepackage{amssymb}
\graphicspath{ {./images/} }
\usepackage{orcidlink}
\usepackage{breqn}
\usepackage{physics}

\usepackage{tikz} 

\usepackage{float}

\usepackage{multirow}
\usepackage{diagbox}

\usepackage{graphicx}  

\usepackage[T1]{fontenc}

\usepackage{threeparttable}  

\usepackage{pifont}

\usepackage{bbding}
\usepackage{wasysym}

\usepackage{cleveref}

\usepackage{makecell} 

\usepackage{bm}

\newif\ifshowmods
\showmodsfalse   

\ifshowmods
                 
                \newcommand{\Ablue}[1]{\textcolor{blue}{#1}}
                 
\else
                
                \newcommand{\Ablue}[1]{#1}
                
\fi

\begin{document}

\title{
Adversarial Attacks on Deep Learning-Based False Data Injection Detection in Differential Relays
}

\author{
Ahmad~Mohammad~Saber\orcidlink{0000-0003-3115-2384},~\IEEEmembership{Member,~IEEE,}
        Aditi~Maheshwari\orcidlink{0009-0009-2959-3552},
        Amr~Youssef\orcidlink{0000-0002-4284-8646},~\IEEEmembership{Senior Member,~IEEE,}
        and~Deepa~Kundur\orcidlink{0000-0001-5999-1847},~\IEEEmembership{Fellow,~IEEE}
\thanks{Ahmad Mohammad Saber, Aditi Maheshwari, and Deepa Kundur are with 
the Department of Electrical and Computer Engineering, University of Toronto, Toronto, ON M5S 1A1, Canada
(e-mails: \href{mailto:ahmad.m.saber@ieee.org}{ahmad.m.saber@ieee.org};
\href{mailto:aditi.maheshwari@mail.utoronto.ca}{aditi.maheshwari@mail.utoronto.ca};
\href{mailto:dkundur@ece.utoronto.ca}{dkundur@ece.utoronto.ca}).
Amr Youssef is with the Concordia Institute for Information Systems Engineering (CIISE), Concordia University, Montreal, QC, Canada
(e-mail: 
\href{mailto:youssef@ciise.concordia.ca}{youssef@ciise.concordia.ca}).}
}

\markboth{
}%
{Shell \MakeLowercase{\textit{et al.}}: Bare Demo of IEEEtran.cls for IEEE Journals}

\maketitle

\begin{abstract}
The application of Deep Learning-based Schemes (DLSs) for detecting False Data Injection Attacks (FDIAs) in smart grids has attracted significant attention. This paper demonstrates that adversarial attacks—carefully crafted FDIAs—can evade existing DLSs used for FDIA detection in Line Current Differential Relays (LCDRs). We propose a novel adversarial attack framework, utilizing the Fast Gradient Sign Method, which exploits DLS vulnerabilities by introducing small perturbations to LCDR remote measurements, leading to misclassification of the FDIA as a legitimate fault while also triggering the LCDR to trip. We evaluate the robustness of multiple deep learning models—including multi-layer perceptrons, convolutional neural networks, long short-term memory networks, and residual networks—under adversarial conditions. 
Our experimental results demonstrate that while these models perform well, they exhibit high degrees of vulnerability to adversarial attacks. For some models, the adversarial attack success rate exceeds 99.7\%.
To address this threat, we introduce adversarial training as a proactive defense mechanism, significantly enhancing the models' ability to withstand adversarial FDIAs without compromising fault detection accuracy. Our results highlight the significant threat posed by adversarial attacks to DLS-based FDIA detection,  underscore the necessity for robust cybersecurity measures in smart grids, and demonstrate the effectiveness of adversarial training in enhancing model robustness against adversarial FDIAs.

\end{abstract}


\IEEEpeerreviewmaketitle

\section{Introduction}\label{section:Introduction}

\IEEEPARstart{S}{mart}  technologies have significantly improved power system efficiency, reliability,  and adaptability.
However, as grid operations become more digitized and connected,  the risk of cyberattacks, such as false data injection attacks (FDIAs) in which malicious data are injected to manipulate grid protection schemes, increases \cite{Reiter_FDIA}. The devastating impacts of such cyberattacks have made energy systems, including smart grids, attractive targets for malicious cyber attackers \cite{manias2024trends}.

A crucial component of these smart grids is the Line Current Differential Relay (LCDR), which is vital for protecting critical transmission lines by detecting and isolating faults to prevent the risk of cascading failures \cite{howLCDRworks}. LCDRs are favored for line protection when speed and sensitivity are crucial, such as in microgrids, which are themselves prime targets of cyberattacks \cite{NREL_military_MGs, karanfil2022detection}. 
 Unlike other protective relays, such as directional overcurrent and distance relays, LCDRs are more sensitive to faults, even those of high impedance, and are less affected by issues arising from the presence of distributed generators within microgrids \cite{MG_protection}.
 However, the full reliance of LCDRs on unsecured communication exposes LCDRs to significant cyber threats, particularly FDIAs \cite{Ahmad_smartgrid}.
FDIAs pose a severe threat by manipulating remote current measurements exchanged between LCDRs, potentially leading to false tripping of lines and unnecessary relay operations \cite{Remedial_pilot_main_protection}. The infamous cyberattacks on the Ukrainian power grid in December 2015, where coordinated attacks on SCADA systems resulted in power outages for more than 230,000 customers, underscore the critical need for robust protection mechanisms that can withstand such cyber threats and ensure grid operation resilience \cite{case2016analysis}.

To counter these threats, machine learning (ML) and deep learning (DL) algorithms have been proposed to detect FDIAs within smart grid environments. In LCDRs, the goal of these schemes is to prevent unnecessary tripping of the LCDR caused by FDIAs that manipulate the attacked LCDR's remote measurements to satisfy its operating criterion similar to a genuine fault. As depicted in Algorithm \ref{alg:DL_FDIA_Detection_in_LCDRS},  to detect such FDIAs, once the LCDR's fault-detection module is triggered to trip, the FDIA-detection module, which employs a DL model, promptly distinguishes between genuine faults (that should be detected by the LCDR) and false-tripping FDIAs \cite{mypaper_TII}.
Although these traditional FDIAs can often be detected by deep learning-based FDIA detection schemes (DLSs) such as those in LCDRs,
adversarial attacks—a stealthier form of FDIA—can evade DL-based detection systems \cite{goodfellow2014explaining}.

\begin{algorithm}[t!]
\caption{DL-based Detection of FDIAs in LCDRs}
\label{alg:DL_FDIA_Detection_in_LCDRS}
\SetNoFillComment
\KwIn{LCDR's local and remote measurements}
\KwOut{
    \textit{FDIA Alarm}: 0 or 1 \tcp*[l]{1: Alarm activated (FDIA detected)}\\ 
    \>\>\>\>\>\>\>\>\>\textit{Trip Command}: 0 or 1 \tcp*[l]{ 1: send a \\ signal to trip the protected line}
}
\If{LCDR's \textbf{fault detection module} is triggered to trip \tcp*[l]{could be a fault or an FDIA}}{
    Activate the \textbf{DL}-based \textbf{\textit{FDIA detection module}} \tcp*[l]{DL model predicts whether LCDR measurements are of a fault or an FDIA}
    \If{\textbf{FDIA detection module} detects an FDIA}{
        set \textit{FDIA Alarm} $\gets$ \textbf{1}; \textit{Trip Command} $\gets$ \textbf{0};
    }
    \Else{
        set \textit{FDIA Alarm} $\gets$ \textbf{0}; 
        \textit{Trip Command} $\gets$ \textbf{1};
    }
    \Return{\textit{FDIA Alarm}, \textit{Trip Command}}
}
\end{algorithm}

\begin{figure}[t!]
\centering
\includegraphics[width=1\columnwidth]{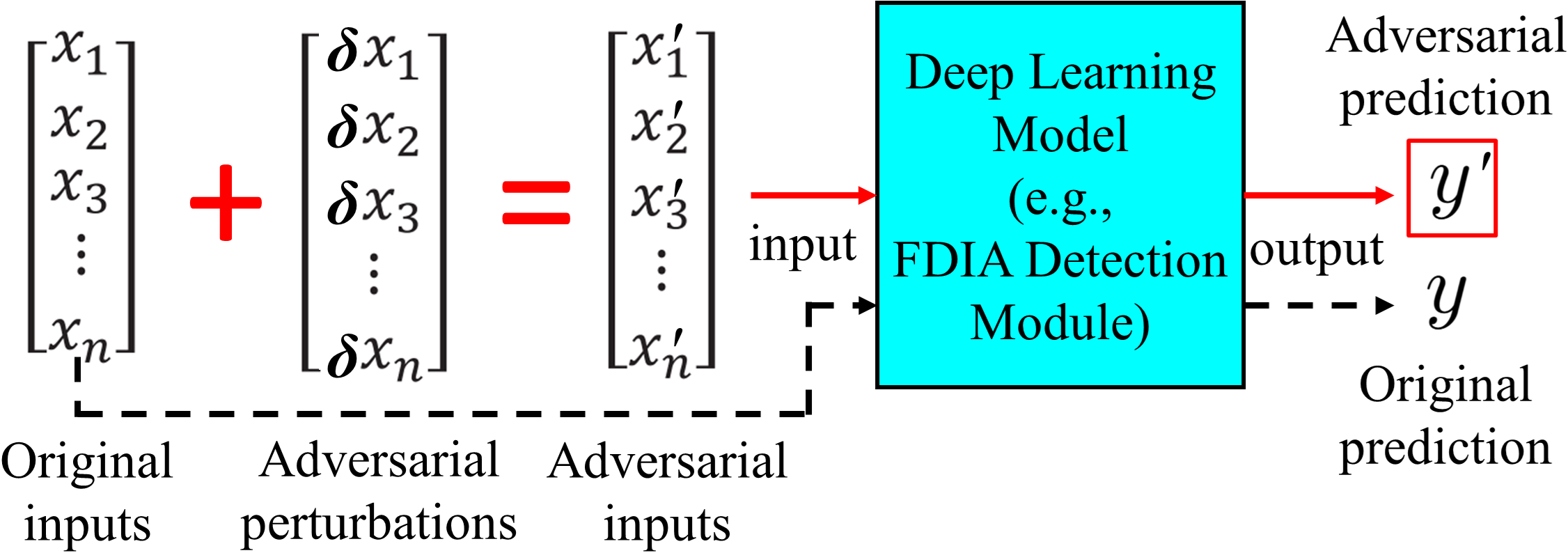}
\caption{Adversarial attack process: Adding subtle perturbations to input measurements causes DL model misclassification.}
\label{fig:Adversarial_Illustration}
\end{figure}

Adversarial attacks, such as those generated by the Fast Gradient Sign Method (FGSM) \cite{andriushchenko2020understanding,carlini2017towards}, are crafted FDIAs designed to deceive DL models \cite{goodfellow2014explaining,ren2020adversarial,DLLimitations}. These attacks directly target and exploit vulnerabilities in DL models used in smart grid components by introducing subtle perturbations to the original input, leading to misclassification \cite{sanchez2024attacking,chakraborty2018adversarial,zhang2024vulnerability}. An illustration of these attacks is shown in Fig. \ref{fig:Adversarial_Illustration}. 
The ability of adversarial attacks to deceive DL-based detection mechanisms poses a significant challenge to secure critical smart grid components that are dependent on DL models, such as LCDRs. 
Adversarial FDIAs against LCDRs can mislead FDIA-detection DLSs into classifying manipulated measurements as legitimate faults, resulting in unnecessary relay tripping and potentially jeopardizing power system stability \cite{mypaper_TII}. 
Furthermore, the stealthiness of adversarial attacks can erode trust in DL-based detection systems for smart grids, especially as these systems evolve and gain adoption. This loss of confidence can undermine the reliability and benefits of such systems in the competitive smart grid landscape. Undermining trust may be a key motive behind these attacks, affecting both targeted systems and the broader acceptance of machine learning solutions. Therefore, investigating the impact of adversarial attacks on DL-based FDIA detection systems is crucial to ensure security and maintain confidence in DL approaches in smart grids.

In this respect, existing research falls into two main categories: securing LCDRs against FDIAs and adversarial attacks targeting smart grid components other than LCDRs.
The first category focused on cybersecurity solutions for LCDRs. Early approaches, such as enhancing resilience through measurement-based detection systems 
\cite{Remedial_pilot_main_protection},
faced limitations due to the reliance on complex system estimations and the vulnerability to sophisticated FDIAs. Model-based methods, often applied to microgrid settings \cite{ResMVDC, Cyber_Resilient_Protection}, also struggled with accuracy and system complexity, revealing a need for model-free and secure alternatives.
\Ablue{Two defense strategies against man-in-the-middle and  relay setting attacks were proposed in \cite{saber2024unmasking} and \cite{ganjkhani2022optimal}, respectively.}
Recent studies have turned to DL for FDIA detection, with methods such as \Ablue{multi-layer perceptrons (MLPs)} and autoencoders showing promise \cite{mypaper_TII}. However, these models remain vulnerable to adversarial attacks specifically crafted to evade detection, a challenge that previous works have not addressed.
\Ablue{The second category, summarized in Table \ref{tab:adversarial_attacks_related}, have primarily focused on adversarial attacks targeting other smart grid components, e.g., load forecasting systems \cite{chen2019exploiting}, state estimation \cite{sayghe2020evasion,li2021towards},
power quality recognition systems \cite{khan2024adaptedge,tian2021adversarial}, power system stability assessment \cite{zhang2023cybersecurity}, voltage and transient stability assessment schemes \cite{ren2022robustness,ren2022vulnerability}, renewable energy output forecasting \cite{ruan2024vulnerability}, inertial forecasting \cite{chen2023vulnerability}, grid event classification \cite{niazazari2020attack}, and optimal power flow control \cite{zeng2023physics}.}
However, none have considered the dual challenge of attacking both DL-based detection models and triggering the physical relay operation, as is required for attacks on LCDRs. 

\begin{table}[t!]
\centering
\caption{\Ablue{Summary of related works on adversarial attacks against smart grid components}}
\begin{tabular}{c c c c}
\Xhline{3\arrayrulewidth}
\rule{0pt}{3ex}  \textbf{Ref}
&
 \textbf{Targeted Smart Grid Function}
& 
\textbf{Attack}
& 
\textbf{Defense}
\\  \hline
\rule{0pt}{3ex} \cite{chen2019exploiting} &  Load forecasting systems & \includegraphics[height=0.25cm]{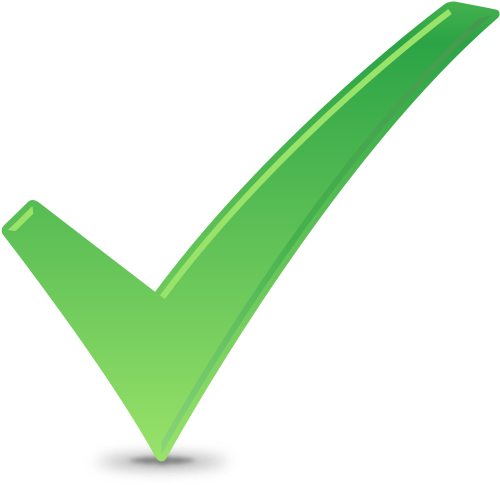}   & \includegraphics[height=0.25cm]{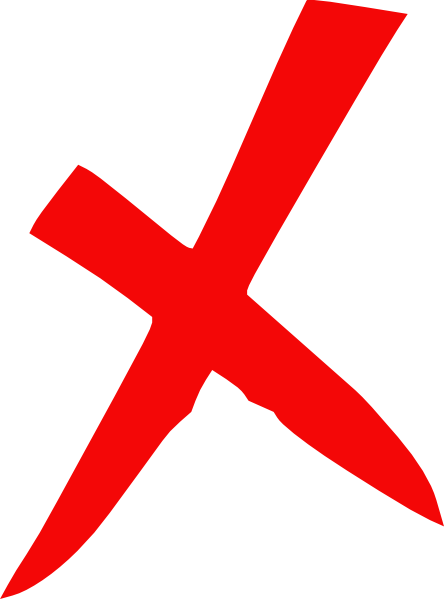} \\
\rule{0pt}{3ex} \cite{sayghe2020evasion,sayghe2020adversarial} & State estimation in power systems & \includegraphics[height=0.25cm]{tick.png}   & \includegraphics[height=0.25cm]{cross} \\
\rule{0pt}{3ex} \cite{li2021towards,tian2022adversarial} & State estimation in power systems & \includegraphics[height=0.25cm]{tick.png}   & \includegraphics[height=0.25cm]{tick.png} \\
\rule{0pt}{3ex} \cite{khan2024adaptedge} & Power quality recognition systems & \includegraphics[height=0.25cm]{tick.png}   & \includegraphics[height=0.25cm]{cross.png} \\
\rule{0pt}{3ex} \cite{tian2021adversarial} & Power quality recognition systems & \includegraphics[height=0.25cm]{tick.png}   & \includegraphics[height=0.25cm]{tick.png} \\
\rule{0pt}{3ex} \cite{zhang2023cybersecurity} & Power system stability assessment & \includegraphics[height=0.25cm]{tick.png}   & \includegraphics[height=0.25cm]{cross.png} \\
\rule{0pt}{3ex} \cite{ren2022robustness} & 
Stability assessment schemes & \includegraphics[height=0.25cm]{tick.png}   & \includegraphics[height=0.25cm]{cross.png} \\
\rule{0pt}{3ex} \cite{ren2022vulnerability} & 
Stability assessment schemes & \includegraphics[height=0.25cm]{tick.png}   & \includegraphics[height=0.25cm]{tick.png} \\
\rule{0pt}{3ex} \cite{ruan2024vulnerability} & Renewable energy output forecasting & \includegraphics[height=0.25cm]{tick.png}   & \includegraphics[height=0.25cm]{cross.png} \\
\rule{0pt}{3ex} \cite{chen2023vulnerability} & Inertial forecasting & \includegraphics[height=0.25cm]{tick.png}   & \includegraphics[height=0.25cm]{cross.png} \\
\rule{0pt}{3ex} \cite{niazazari2020attack} & Grid event classification & \includegraphics[height=0.25cm]{tick.png}   & \includegraphics[height=0.25cm]{tick.png} \\
\rule{0pt}{3ex} \cite{zeng2023physics} & Optimal power flow control & \includegraphics[height=0.25cm]{tick.png}   & \includegraphics[height=0.25cm]{cross.png} \\
\hline
\rule{0pt}{3ex} This work & LCDRs & \includegraphics[height=0.25cm]{tick.png}  & \includegraphics[height=0.25cm]{tick.png}  \\ 
\Xhline{3\arrayrulewidth}
\end{tabular}
\label{tab:adversarial_attacks_related}
\end{table}

To our knowledge, no prior work investigated the vulnerabilities of DL-based FDIA detection systems in LCDRs against adversarial attacks, despite the critical role LCDRs play in 
line protection.
This problem also encompasses a unique additional set of objectives and constraints that must be taken into consideration to design successful adversarial attacks against the LCDR.
For instance, for an adversarial attack to succeed, it must not only deceive the DLS but also trigger the LCDR to trip, adding complexity beyond scenarios where decision-making relies solely on a machine-learning model.
Herein, the adversarial samples must be misclassified by the DLS as faults, since the primary objective of the attacker is to cause the LCDR to trip unnecessarily in the absence of a real fault.
Moreover, the problem is constrained by the requirement that only features from remote measurements can be manipulated, while local measurement features remain unchanged. Local measurements, being closely tied to the relay, are difficult to manipulate as they are transmitted directly through copper wires, whereas remote measurements, which traverse longer distances and potentially vulnerable media, offer a broader attack surface. This distinction highlights the need for robust detection systems capable of withstanding targeted adversarial attacks. Finally, for LCDRs, these robust detection systems must not negatively impact the LCDR's ability to detect actual faults.
Therefore there is still a persistent gap to evaluate the impact of adversarial attacks on LCDRs and develop an adequate defense mechanism.

This paper addresses this crucial gap in the literature by unveiling and investigating the vulnerability of DL models used in LCDRs for detecting FDIAs to adversarial attacks and proposing a proactive defense mechanism.
In this regard, the key contributions of this paper are:
\begin{itemize}
    \item 
    We develop a novel adversarial attack framework focused specifically on DLSs designed for FDIA detection in LCDRs, which play a critical role in
    line protection. A key aspect of this framework is the focus on remote measurements, which are vulnerable due to long-distance communication, while local measurements remain secure. This selective manipulation presents an additional challenge for adversarial attack generation, as it restricts the attack space compared to traditional scenarios where all features are accessible for perturbation. Additionally, the adversarial attack must meet the dual success criteria of misclassifying FDIAs as faults while at the same time triggering the LCDR to trip, a unique requirement not addressed in prior adversarial attack studies on smart grid components. Adversarial attacks targeting such a combination of objectives have not been studied in related work on smart grid cybersecurity. 
    \item 
    We conduct an extensive evaluation of multiple DLS models under adversarial attack conditions in the context of LCDRs, which is crucial for assessing model-specific vulnerabilities. 
    These models include Multi-Layer Perceptrons (MLP), proposed in the most recent LCDRs research papers, Convolutional Neural Network (CNN), Long Short-Term Memory (LSTM), and Residual Neural Network (ResNet) models. Each model is trained, before evaluation, on a comprehensive dataset of a) FDIAs with different attack vectors, and b) faults of varying types, locations, impedances, and inception angles. The results reveal
    valuable new insights that while these DLS models can differentiate between naive FDIAs and faults,  all models are susceptible to a degree to adversarial attacks, with models such as MLP, exhibiting higher vulnerability.
    \item 
    We introduce adversarial training as a defense mechanism tailored specifically for DL-based FDIA detection in LCDRs, an application that has not been explored in previous works. While adversarial training has been used in other domains, this is the first study to apply it in the context of smart grid protection. Our results demonstrate that adversarial training significantly improves the models' ability to detect  FDIAs while preserving the LCDR's ability to accurately detect legitimate faults. 
\end{itemize}
\section{Preliminaries}\label{section:Preliminaries}
\subsection{Vulnerability of LCDRs to False-Data Injection Attacks}\label{section:LCDRs_Vulnerability}
\begin{figure}[t!]
\centering
\includegraphics[width=0.5\columnwidth]{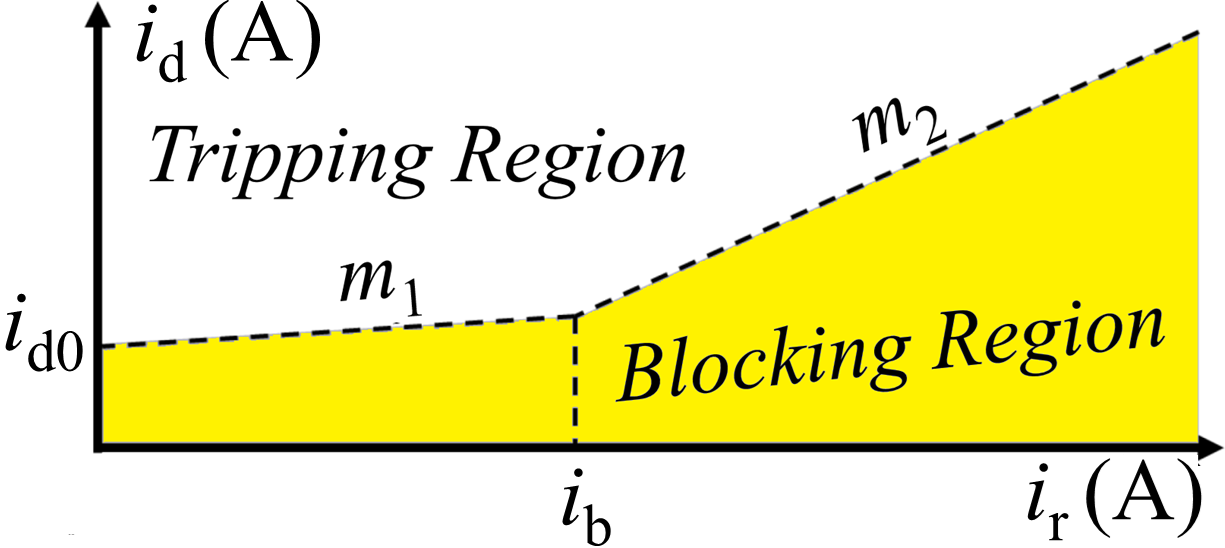}
\caption{Operating characteristics of the LCDR.}
\label{fig:LCDR_ccs}
\end{figure}

To protect a line, an LCDR, located at one end of the line, receives both the local current measurements (measured at the line terminal close to the LCDR) and the remote current measurements (communicated from the other side of the line)  \cite{howLCDRworks}. At the  $t$-th time step, the differential current ($\mathbf{i_d}$) and restraining current ($i_r$) are calculated as:
\begin{equation}\label{eqn:differential}
    \mathbf{i_d} [t] =  \mathbf{i}_1[t] + \mathbf{i}_2[t]  
\end{equation}  

\begin{equation}
    i_r [t] = || \mathbf{i}_1[t] || + || \mathbf{i}_2[t] ||
\end{equation}
\noindent where $\mathbf{i}_1 [t]$ and $\mathbf{i}_2 [t]$ are the local and remote current samples, respectively, at time step $t$, and $||\cdot||$ denotes the magnitude of the enclosed vector. Afterward, the operating current $\mathbf{i}_{op}$ is determined as:
\begin{equation}
\mathbf{i}_{op} [t] =                       
\begin{cases}  
   i_{d0} + m_1 \times i_r [t]   & \text{if } i_r [t] \leq i_b   \\
   i_{d0} + m_1 \times i_b + m_2 (i_r [t] - i_b) & \text{if } i_r [t] > i_b 
\end{cases}
\end{equation}
\noindent where $i_{d0}$, $i_b$, $m_1$, and $m_2$ are the LCDR's operational settings, as depicted in Fig. \ref{fig:LCDR_ccs}. The LCDR issues a trip signal when:
\begin{equation}\label{eqn:criterion}
     i_d [t] = || \mathbf{i_d} [t] ||      \ge \mathbf{i}_{op} [t]
\end{equation}
\noindent which is the condition that indicates an internal fault. This is since $\mathbf{i}_1$ and $\mathbf{i}_2$ show significant disparity under fault conditions, but they have approximately the same magnitude and opposite direction otherwise (before considering cyberattacks) \cite{saadat1999power}. 
\begin{figure}[t!]
\centering
\includegraphics[width=1\columnwidth]{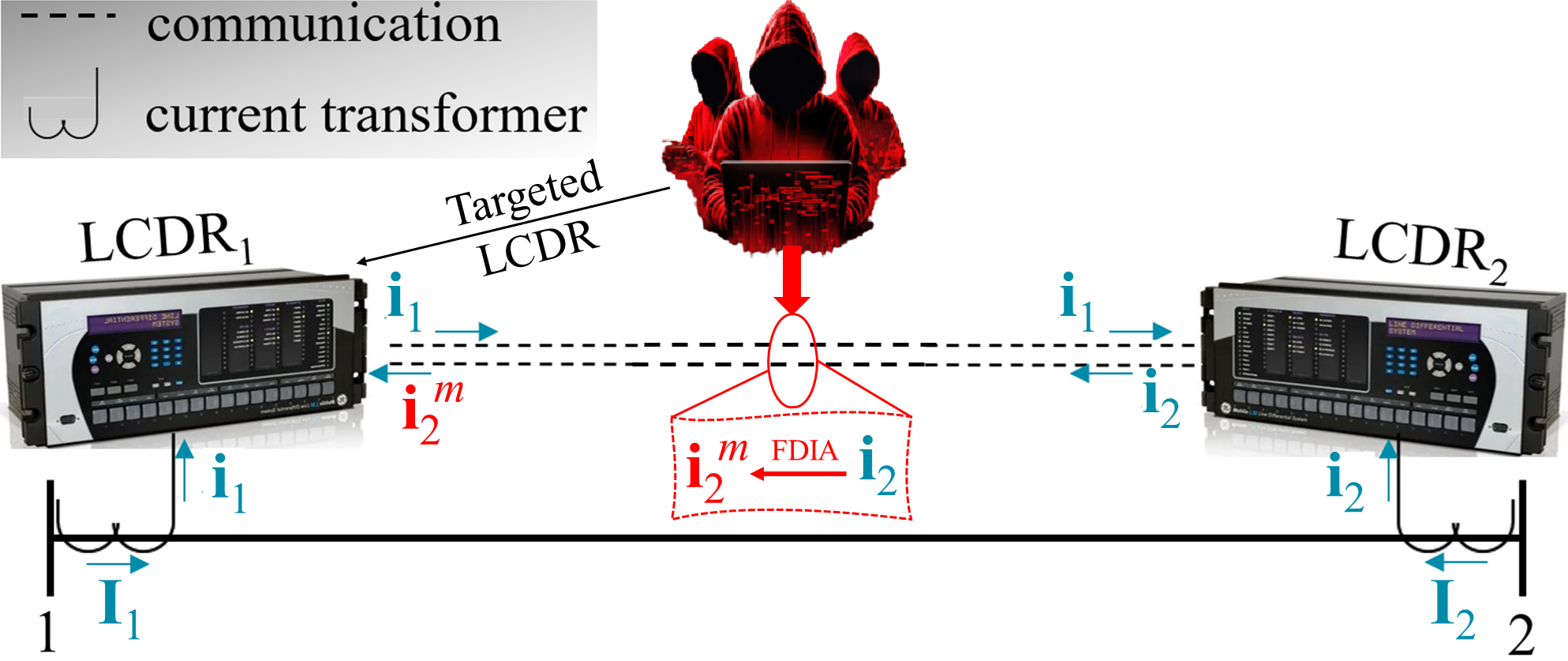}
\caption{Illustration of an FDIA  to falsely trip LCDR$_1$'s line.}
\label{fig:LCDR_idea}
\end{figure}
A malicious entity, targeting a specific line protected by LCDRs, could compromise the communication channel transmitting $\mathbf{i}_2$ and manipulate $\mathbf{i}_2$, as illustrated in Fig. \ref{fig:LCDR_idea} \cite{mypaper_TII}. Such manipulation by a malicious entity with  knowledge of protection principles can shift the operating point of the relay into the trip region, causing a false trip of the line protected by the attacked LCDR.
To achieve this, adversaries can compromise the two-way communication link over which $\mathbf{i}_2$ is transmitted, which is possible since LCDRs often use vulnerable communication or exploit some of the potential weak points, e.g., routers and switches, in the communication link carrying the remote measurements \cite{Ahmad_smartgrid,ResMVDC, mypaper_TII}.  

\Ablue{LCDRs support a variety of communication networks, which can be 
multiplexed communication or direct connections \cite{SEL_inc2}, with a range of communication media, including  wireless communication such radio or microwave \cite{ResMVDC,Cyber_Resilient_Protection} and fiber-optic wires \cite{SEL_inc2}.
Attackers can intrude into the LCDR's communication network over which the remote measurements are communicated. This intrusion can exploit the vulnerability of communication media, e.g., in case of wireless communication. Additionally, attackers can intrude into the LCDR's network, for example in case of fiber optical networks, through  
electrical-to-optical converters, monitoring ports, communication routers, repeaters and other intermediary devices, which can be exploited by malicious entities and used for eavesdropping and injection attacks \cite{opticfiber, opticfiber2}. 
Furthermore, irrespective of the type of communication, attackers can  exploit the vulnerabilities of the IEC 61850 standard, based on which modern LCDRs communicate and exchange time-stamped instantaneous measurements as sampled value packets \cite{SEL_inc2,yang2015cybersecurity},
to intrude into LCDR's wide area  or local area networks, e.g., through access points or compromised devices connected to the communication network, allowing them to manipulate the 
LCDR sampled values measurements  
exchanged between protection devices 
\cite{yang2015cybersecurity,chattopadhyay2017toward,john2013tcp}. }

Once the communication link is breached, attackers can intercept, inject, replace, or manipulate the LCDR's remote measurements \cite{Dolev, Reiter_FDIA}. Local measurements, however, are much harder to manipulate, and thus considered secure, since they can be directly sent over hard wires from the local current transformers to the LCDR with no room for manipulation.
Based on the above, FDIAs on LCDRs can be modeled as:
\begin{equation}\label{eqn:differential_attack}
  \mathbf{i_d}^m [t] = || \mathbf{i}_1[t] + \mathbf{i}_2^m [t] ||
\end{equation}
\noindent where $\mathbf{i}_2^m$ represents the manipulated $\mathbf{i}_2$ and $\mathbf{i_d}^m$ represents the value of $\mathbf{i_d}$ under $\mathbf{i}_2^m$. Under normal operation of the line, $\mathbf{i_d} [t] = 0$, so (\ref{eqn:differential}) can be rewritten  as:
\begin{equation}
    \mathbf{i}_1[t] = - \mathbf{i}_2[t]
\end{equation}
\noindent Substituting this into (\ref{eqn:differential_attack}) yields:
\begin{equation}
    i_d^m [t] = || -\mathbf{i}_2[t] + \mathbf{i}_2^m [t] ||
\end{equation}
Given that adversaries can eavesdrop on $\mathbf{i}_2$, they can modify $\mathbf{i}_2^m [t]$ to satisfy the relay’s trip condition (\ref{eqn:criterion}).
\Ablue{For example, multiplying original $\mathbf{i}_2$ by -1, which is equivalent to reversing the phase angle of remote measurements, satisfies the LCDR's tripping condition. In this example, the LCDR views the (manipulated) remote measurements as having a similar direction to that of $\mathbf{i}_1$, as in internal faults
 \cite{saadat1999power}, causing the LCDR to trip.}
%
%
%
 %
%
%
%
\subsection{Deep-Learning-Based Detection of FDIAs on LCDRs}\label{section:FDIA_Detection}
The use of DL models for detecting FDIAs against LCDRs builds on the ability of such models to classify multivariate time-series data with high accuracy \cite{ismail2019deep,lecun2015deep}. After the LCDR is triggered by either a fault or an FDIA at time step $k$, the LCDR measurements vector, containing the LCDR measurements recorded for a few milliseconds before and after the LCDR is triggered, can be represented as  \( \mathbf{x}  [t] \in \mathbb{R}^{ d \times T}\), where \(d\) denotes the dimension of the input features, corresponding to the 3-phase local and remote measurements of the LCDR, and \(T\) is the number of time steps (or sequence length). The goal is to train a model \(\mathbf{f}_{\bm{\theta}}\), parameterized by  $\bm{\theta}$, that can distinguish between genuine faults and FDIAs. Once trained, this model should require LCDR measurements vector $\mathbf{x}$ as the only input.
The prediction made by such a model is denoted $\hat{y} \in \{0, 1\}$, such that  \(\hat{y} = 0\) and  \(\hat{y} = 1\) indicate that the model predicts the triggering event as a legitimate fault and as a malicious FDIA, respectively. The model is trained offline using a training dataset ($\mathbf{D}_{train}$) consisting of $N$ samples. 
Each sample consists of the $d$-dimensional input time series \(\mathbf{x}\) and its corresponding label \(y \in \{0, 1\}\), where \(y = 0\)  and \(y = 1\) denote an FDIA and a fault, respectively. The objective during training is to minimize the binary cross-entropy loss function \(J(\hat{y}, y)\) \cite{lecun2015deep}:
\begin{equation}
J = -\frac{1}{N} \sum_{i=1}^{N} \left[ y_i \log(\hat{y}_i) + (1 - y_i) \log(1 - \hat{y}_i) \right] 
\end{equation}
Training is performed iteratively over a number of epochs using the backpropagation concept along with an optimization method such as stochastic gradient descent \cite{lecun2015deep}.
While training, the DL model \(\mathbf{f}_{\bm{\theta}}\) learns complex patterns in LCDR measurements in both fault cases and FDIA cases.
Once the DL model is trained, it can be used in the LCDR to detect FDIAs as soon as they trigger the LCDR.
That is, once the LCDR's fault-detection module is triggered to trip due to an event (a fault or an FDIA), the trained model would use the LCDR's measurements to predict whether the event is a legitimate fault or a malicious FDIA. The LCDR will trip only if the model confirms the triggering event is a fault, ensuring that the LCDR does not trip due to an FDIA.
Various DL models can be employed for FDIA detection, each with unique architectures for processing the LCDR measurements \cite{lecun2015deep, alzubaidi2021review}. In this paper, we investigate the following deep neural networks-based models: 
2) CNNs, which apply spatial filters to the input data, capturing local patterns across time steps \cite{alzubaidi2021review},
3)  LSTM  networks, which are designed to capture temporal dependencies in sequential data, making them suitable for time-series analysis \cite{alzubaidi2021review}, and
4) ResNets, which introduce skip connections between layers, allowing for deeper architectures by mitigating the vanishing gradient problem \cite{alzubaidi2021review, lecun2015deep}.
Although these DL-based schemes are expected to perform well in detecting FDIAs in general, 
they can be bypassed by more stealthy adversarial FDIAs, as discussed in the next Section.
\section{Adversarial Attacks Against  DL-based FDIA-detection Schemes in LCDRs %
}\label{section:Adversarial_Attack_Model}
%
 %
\subsection{Threat Model} \label{section:threat_model}
Before deploying DL-based FDIA detection systems for operation in real-world power systems, it is essential to evaluate their robustness against potential adversarial FDIAs. Utilities should conduct penetration tests where malicious entities simulate such attacks, providing empirical evidence of how well detection systems can withstand adversarial manipulation and helping power system operators prepare effective countermeasures. The scenarios outlined here represent a worst-case threat model designed to explore vulnerabilities under the most challenging conditions, allowing utilities and DL vendors to establish upper bounds for the performance degradation of DL-based FDIA detection models under adversarial attacks.
\subsubsection{Attack Objective}
We assume an adversary whose objective is to manipulate the remote current measurements of an LCDR, bypasses the DLSs for FDIA detection, and causes the LCDR to falsely trip, disconnecting a protected line.
\subsubsection{Attacker Capabilities Regarding the Network}

We follow the \emph{Dolev–Yao} threat model, where attackers control the communication link used by the targeted LCDR, enabling them to eavesdrop on, manipulate, and inject messages (measurements) exchanged between LCDRs \cite{Dolev}. The attacker can intercept and manipulate both the magnitude and phase angle of the remote current measurements ($\mathbf{i}_2$) \cite{Dolev}, which are transmitted over potentially vulnerable channels \cite{Remedial_pilot_main_protection}. 
Furthermore, the attacker is assumed to understand the LCDR’s protection logic, including equations (\ref{eqn:differential})-(\ref{eqn:criterion}). This knowledge allows attackers to craft manipulations that deceive the LCDR into misinterpreting remote measurements as faults. 
However, local current measurements cannot be manipulated, as they are transmitted securely over copper wires to the LCDR. Thus, attacks are confined to remote measurements, which are more vulnerable due to their reliance on less secure communication media. Additionally, attackers cannot directly create physical faults or disconnect parts of the power system; otherwise, they could disconnect the targeted line without a cyberattack.

\subsubsection{Attacker Capabilities Regarding the DL Model}

The attackers are presumed capable of acquiring the FDIA detection model or obtaining crucial information to replicate it through stolen credentials and malware attacks. They may collaborate with compromised insiders or be involved in the DL model’s training, either as a utility company member or as part of the organization supplying the model \cite{NISTGuidelines}. This involvement grants access to the model’s architecture, parameters, and training data, enabling the creation of sophisticated adversarial perturbations.
The attacker is also assumed to have access to a subset of FDIA samples used to train the DLS. These samples, \Ablue{can be} generated to meet basic differential protection criteria without considering FDIA detection, serve as a foundation for adversarial attacks. Alternatively, FDIA samples can be generated using publicly available information on power system protection principles and commercial LCDR catalogs. For example, the attacker can create FDIA samples by multiplying $\mathbf{i}_2$ of the targeted LCDR by random negative integers that satisfy (\ref{eqn:differential})-(\ref{eqn:criterion}).
The adversary employs algorithms like FGSM (explained in Section \ref{section:FGSM}) to introduce small perturbations to FDIA samples in a black-box manner, i.e., without accessing the exact model parameters. These samples have a high chance of bypassing the DLS and triggering the LCDR to trip.
\subsection{ 
FGSM for Generating Successful Adversarial Attacks Against LCDRs with DL FDIA Detection systems
}\label{section:FGSM}
In this section, we utilize the FGSM, a widely used and computationally efficient adversarial attack algorithm \cite{goodfellow2014explaining}, as a tool to generate adversarial FDIA samples to bypass DL-based FDIA detection systems in LCDRs and trigger the LCDR to trip. 
While other methods can also be used to generate adversarial examples, e.g., projected gradient descent\cite{madry2018towards} and Carlini \& Wagner attacks \cite{carlini2017towards}, FGSM's reliance on the one-step gradient calculation, as will be explained later, offers a balance between attack success and efficiency \cite{goodfellow2014explaining}.
Additionally, FGSM is  
scalable across different DL models and offers controllability on the perturbation magnitude which provides flexibility in balancing attack stealth and effectiveness \cite{ismail2019deep}. This makes FGSM a robust choice for testing the DLS models' robustness.

During the generation of adversarial examples against LCDRs, specific problem constraints must be observed, which are: 
1) perturbations are applied exclusively to features originating from the LCDR's remote measurements since remote measurements must be exchanged over long-distance communication that is often compromised of vulnerable media. Thus, remote measurements are easier to manipulate as they have a wider attack space than local measurements which are directly sent from the current transformers to the LCDR through copper wires, 2) manipulation is restricted only to FDIA samples within the original testing dataset ($\mathbf{D}_{test}$). 
Recall that the attacker's goal is to cause these FDIA samples to be misclassified as faults by the models, causing false tripping of the line protected by the LCDR. Therefore, fault samples in $\mathbf{D}_{test}$ are left unchanged, and 3) dual success criterion: an adversarial sample is considered successful if it deceives not only the individual models but also triggers the LCDR to trip. 

Taking into consideration the above constraints, FGSM is used as a tool to generate adversarial FDIA samples by introducing small, carefully crafted perturbations to the original input sample, labeled as FDIA, with the intent of maximizing prediction errors across the given model, so the DLS model misclassifies this sample as a fault.
For a given model with $\bm{\theta}$ parameters, each adversarial sample ($\mathbf{x}'$) is derived from its corresponding original input ($\mathbf{x}$). An adversarial sample is considered successful if it is misclassified by the model as the target class which can be represented as:
\begin{equation}\label{eqn:adversarial_sample_delta}
  \hat{y} (\mathbf{x}')  =  y_{target} 
\end{equation}
\noindent where  $y_{target}$ is the label of the desired target class, i.e., that of a fault.
To derive $\mathbf{x}'$, the adversarial perturbation, denoted $\bm{\delta}_{\mathbf{x}}$, is computed as:
\begin{equation}\label{eqn:adversarial_sample_delta}
  \bm{\delta}_{\mathbf{x}} = \epsilon \cdot \text{sign}(\nabla_{\mathbf{x}} J(\bm{\theta}, \mathbf{x}', y)) \cdot  a
\end{equation}
\noindent where $\epsilon$  is a parameter that controls the magnitude of the perturbation, \(\nabla_{\mathbf{x}}\) is the gradient of the loss function with respect to the input $\mathbf{x}$, $J(\bm{\theta}, \mathbf{x}, y)$ is the loss function quantifying the discrepancy between the predicted label $\hat{y}_i = f_{\bm{\theta}}(\mathbf{x})$ and the true label $y$,  and  $a$ is a parameter used in the above equation to make the perturbation a function of the input sample's amplitude. To compute $a$, firstly we compute, $\mathbf{x}^{abs}$, which is a vector that contains the absolute value of each element in $\mathbf{x}$, then, $a$ is determined as the maximum value across all elements in $\mathbf{x}^{abs}$.  
Computing the gradient, as described in (\ref{eqn:adversarial_sample_delta}), identifies the direction in the input space that would most significantly increase the loss.
 Hence, this gradient information can be used to perturb the original input $\mathbf{x}$ by moving it in the direction that maximizes the loss, thereby crafting an adversarial example.
 While computing the gradient, only features derived from the LCDR's vulnerable remote measurements are perturbed.
After (\ref{eqn:adversarial_sample_delta}), $\mathbf{x}' $ is computed as:
\begin{equation} \label{eqn:adversarial_sample_simple}
\mathbf{x}' = \mathbf{x} + \bm{\delta}_{\mathbf{x}}
\end{equation}
Further, $\mathbf{x}'_{clip}$, a copy of $\mathbf{x}'$, is generated and then clipped such that no element in $\mathbf{x}'_{clip}$ exceeds the largest element in $\mathbf{x}$, and no element in $\mathbf{x}'_{clip}$ is smaller than the smallest element in $\mathbf{x}$ \cite{goodfellow2014explaining}.
Finally, $\mathbf{x}'$ is updated as follows:
\begin{equation} \label{eqn:adversarial_sample_clipping}
\mathbf{x}' \gets \mathbf{x}'_{clip}
\end{equation}
This process--(\ref{eqn:adversarial_sample_delta}) to (\ref{eqn:adversarial_sample_clipping})--is iteratively applied to each FDIA sample in the original testing dataset $\mathbf{D}_{test}$.
At each iteration, the adversarial example is refined to further increase the likelihood of causing a misclassification by the models, while ensuring that the perturbations remain within acceptable bounds to maintain their subtlety.
The iterative process is terminated when either: 1) $\mathbf{x}'$ is misclassified by the DL model as a fault while triggering the LCDR to trip, and hence is considered a successful adversarial FDIA sample and is returned by the process, or 2) the number of iterations reaches its maximum limit while $\mathbf{x}'$ still cannot fool the classifier or cannot trigger the LCDR. In this case, finding an adversarial sample is considered unsuccessful and the algorithm returns the original $\mathbf{x}$. 
A summary of the proposed adversarial attack framework for DLS-based FDIA detection in LCDRs is presented in Algorithm \ref{alg:adversarial_attack_framework}.
Once the iterative process is complete, the generated successful adversarial samples are compiled into a new testing dataset $\mathbf{D}_{test}'$. 
For FDIA samples from $\mathbf{D}_{test}$ that the FGSM could not generate successful samples for, corresponding original FDIA samples are added to \Ablue{$\mathbf{D}_{test}'$.}
Finally, the unaltered fault samples from $\mathbf{D}_{test}$ are then added to $\mathbf{D}_{test}'$. 
By now, $\mathbf{D}_{test}'$ consists of untampered fault samples and FDIA samples that are primarily adversarial FDIA samples generated using Algorithm \ref{alg:adversarial_attack_framework}. 
$\mathbf{D}_{test}'$ is then used to evaluate the robustness of the LCDR FDIA-detection model, as explained in the upcoming sections. 
%
\begin{algorithm}[t!]
\caption{Proposed adversarial attack framework for FDIA-detection models in LCDRs}
\label{alg:adversarial_attack_framework}
\SetNoFillComment
\KwIn{$\mathbf{f}_{\bm{\theta}}$: DLS model with $\bm{\theta}$ parameters, 
$\mathbf{x}$: LCDR measurement vector, 
$y$: $\mathbf{x}$'s true label, 
$\epsilon$: perturbation magnitude, 
$N_{\text{itr}}$: maximum number of FGSM iterations,
$y_{target}$: target class's label}
\KwOut{$\mathbf{x}'$: successful adversarial sample, \textbf{or} $\mathbf{x}$: original sample (if unsuccessful)}
\SetNlSty{textbf}{}{} 
\textbf{Initialize:} $\mathbf{x}' \gets \mathbf{x}$, $n \gets 0$\;
\If{$y = y_{target}$ \textbf{OR} $\hat{y}(\mathbf{x}) = f_{\bm{\theta}} (\mathbf{x}) = y_{target}$ }{
    \Return{$\mathbf{x}$ \tcp*[l]{$\mathbf{x}$ is already (mis)classified as a fault by the model}}
}
\Else{
    \While{$n \leq N_{\text{itr}}$}{
        Compute and update $\mathbf{x}'$ using (\ref{eqn:adversarial_sample_delta})-(\ref{eqn:adversarial_sample_clipping})\; 

        \textbf{Ensure problem constraints:}\;
        \If{updated $\mathbf{x}'$ satisfies FDIA sample constraints}{
            \Return{$\mathbf{x}'$ \tcp*[l]{successful adv. sample}}  
        }
        \Else{
            \Return{$\mathbf{x}$ \tcp*[l]{could not generate a successful adv. sample}}
        }
        
        \If{$n < N_{\text{itr}}$ }{
            $n \gets n + 1$\;
        }
    }
}
\end{algorithm}
%
%
\section{Experimental Setup and Simulation Results}\label{section:Simulation_Results}
 \subsection{Test System}\label{section:Experimental_Setup}
\begin{figure}[t!]
\centering
\includegraphics
[width=1\columnwidth] {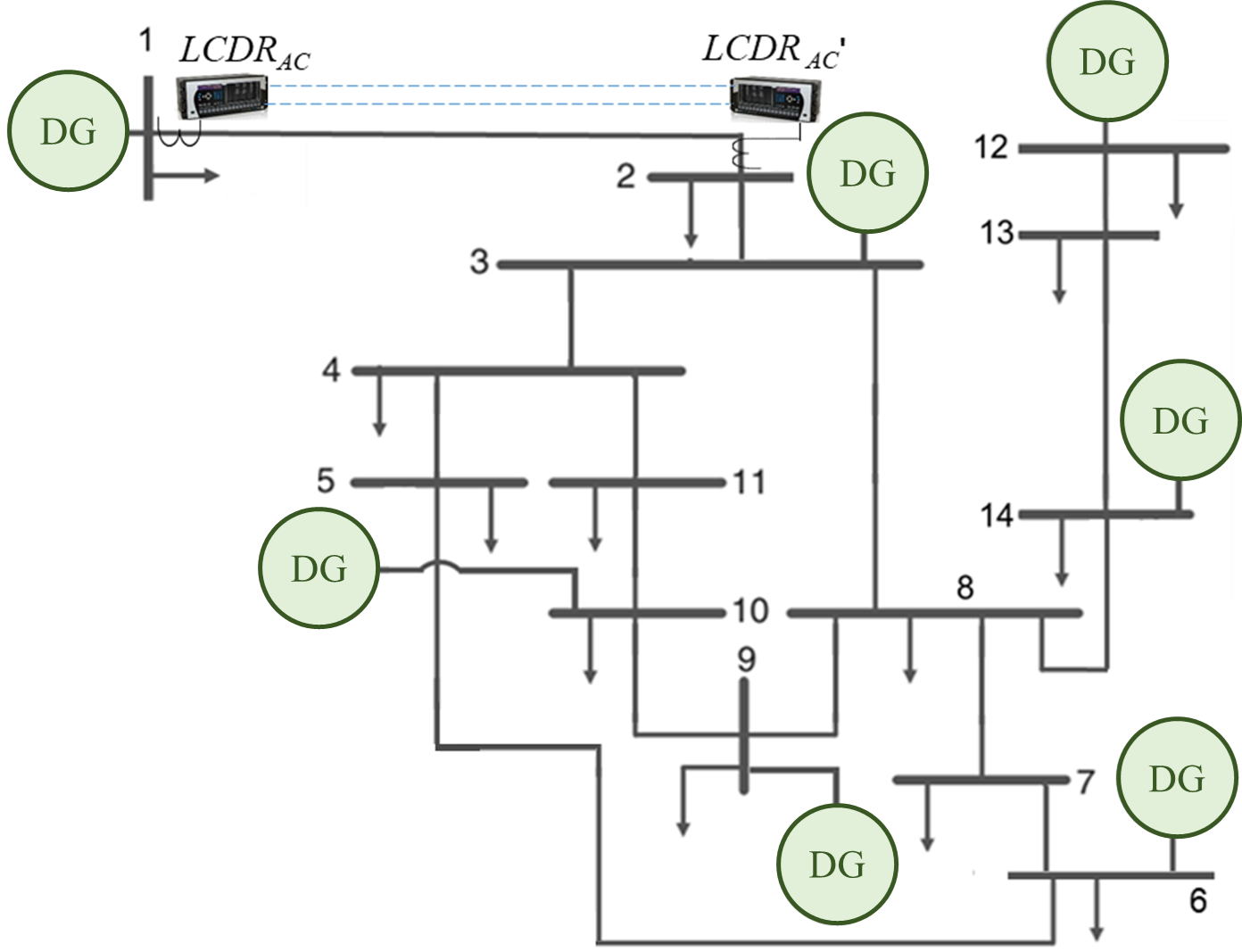}
\caption{ Test system.}
  \label{fig:test_system}
\end{figure}

This section evaluates the impact of adversarial attacks on LCDRs. The evaluation is carried out using the test system shown in Fig. \ref{fig:test_system}, which is simulated in the PSCAD/EMTDC 5.0.2 environment.
The figure depicts a modern inverter-based microgrid that is based on the medium-voltage distribution benchmark network developed by CIGRE
for power system studies. The system's details are available in \cite{rudion2006design}. The test system is supplied by a group of 2-MVA inverter-based generators equipped with 1.5-p.u. fault current limiters to protect the system against excessive fault currents \cite{Saleh}.
In this system, line 1-2 is protected by a pair of LCDRs, denoted $LCDR_{AC}$ and $LCDR_{AC}$\textquotesingle, as illustrated.
$LCDR_{AC}$ is used in this paper as an example to showcase the impact of adversarial attacks on LCDRs.
\Ablue{The LCDR settings $i_{d0}$, $i_b$, $m_1$ and $m_2$ are set as 0.05 kA, 0.585 kA, 0.2 and 0.4, respectively, following \cite{GELCDR,Amir1}, to ensure sensitivity to internal faults while remaining secure against maloperation due to external disturbances.}
Additionally,  measurements sent to the LCDR are combined with additive white Gaussian noise of signal-to-noise ratio randomly varying from 35 to 60 dB to account for the impact of possible measurement noise, which may arise from non-ideal current transformers \cite{schettino2014new}.  

 \subsection{Performance Evaluation Scenarios}\label{section:Dataset}

\begin{figure}[t!]
\centering
\includegraphics[width=0.7\columnwidth]{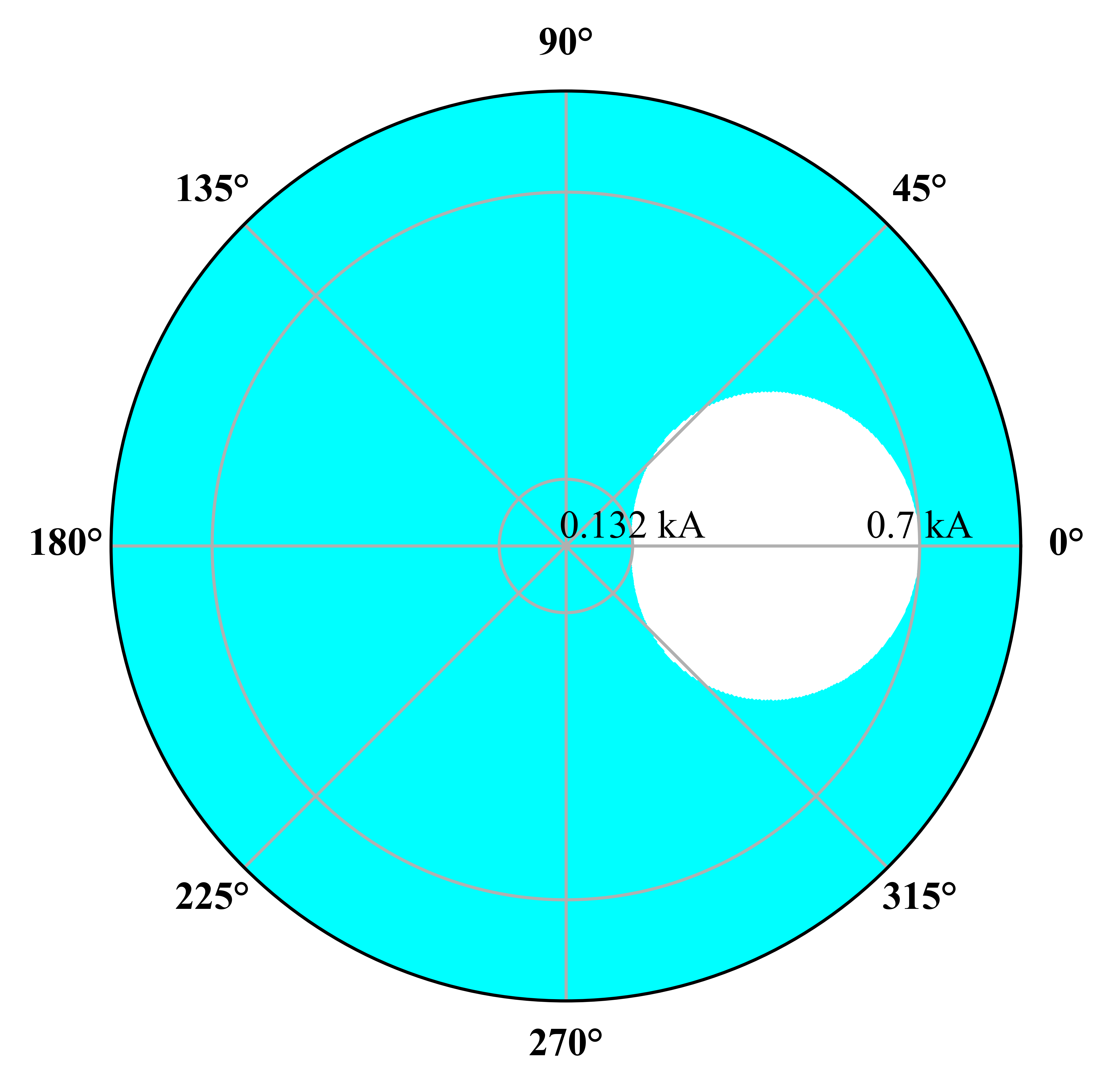}
\\ (a)
\\ \textcolor{white}{.}
\\
\includegraphics[width=0.7\columnwidth]{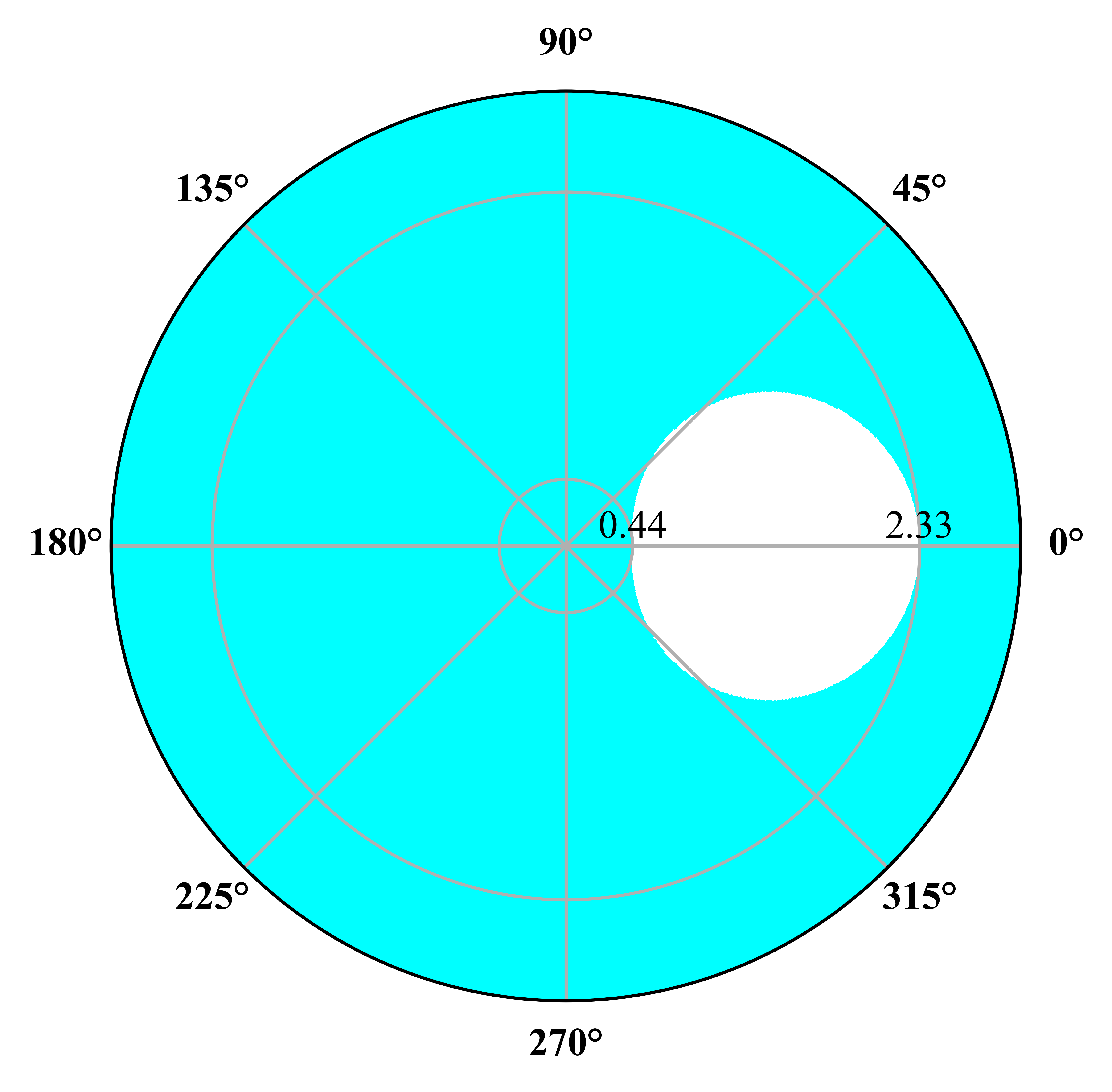}
\\ (b)
\caption{Illustration of $\mathbf{i}_2$ locus for successful FDIAs against $LCDR_{AC}$, (a): $\mathbf{i}_2$ values that would trip $LCDR_{AC}$ (in kA),
(b): 
phasor $\alpha$ values for multiplicative FDIAs ($\mathbf{i}_2 \times \alpha$ yields a value for the remote current that trips the LCDR)
}
\label{fig:vector_inequality_plot}
\end{figure}

The objective of DL-based modules in this paper is to detect FDIAs, by differentiating them from genuine faults, once the LCDR's fault detection module is triggered whether by a fault or an FDIA. For this reason, a comprehensive dataset of internal faults on line 1-2  and FDIAs targeting $LCDR_{AC}$ is generated\footnote{This dataset is available at \url{https://github.com/AhmadMSGit/Data_Adversarial_DL_LCDRs}}.
On the one hand, 
\Ablue{FDIAs are simulated based on the LCDR's operating principle by manipulating the remote measurements of $LCDR_{AC}$ under normal system operation such that the manipulated remote measurements, as received by $LCDR_{AC}$, satisfy the LCDR tripping condition  (1)–(4)  given the LCDR setting values outlined in Section IV.
By numerically solving (1)–(4) under the above conditions, a locus for different $\mathbf{i}_2[t]$ values that  satisfy (1)–(4), as illustrated in Fig. \ref{fig:vector_inequality_plot}, is obtained.
(Herein, FDIAs are simulated considering only the LCDR without considering DL FDIA detectors.)
In our simulations, FDIA samples are generated by intercepting the original $\mathbf{i}_2[t]$ on the onset of the FDIA and 
replacing it with currents whose phasors  (magnitude and angle) values are randomly sampled 
from the light blue locus illustrated in \ref{fig:vector_inequality_plot} a), which is equivalent to multiplying original $\mathbf{i}_2[t]$ by a phasor $\alpha$ whose value is sampled from Fig. \ref{fig:vector_inequality_plot} b).}
All the above is performed under varying load conditions.
In total, 
23,040
FDIA samples are generated. 
\Ablue{For each sample, the attacked LCDR local and remote measurements are recorded for a period of four power cycles, starting two cycles before the attack inception. The observation window spans four 60-Hz power cycles (66.67 ms), capturing three-phase local and remote current measurements in magnitude and phase angle, sampled at 1 kHz. These raw samples are preprocessed with additive white Gaussian noise with 35–60 dB SNR, but undergo no additional filtering or bad data detection, as the focus is solely on FDIA detection rather than distinguishing noise-induced errors.}
On the other hand, simulated fault conditions include 1) Different fault types, including all variations of single-phase, two-phase, and three-phase faults, involving and not involving the ground whenever possible. 
2) Faults under a wide spectrum of impedance values ranging from 0 to 100 $\Omega$, 3) Faults on different locations on the protected line ranging from 10\% of the line (measured from the LCDR's end) to 90\% of the line, and 4) Faults occurring at different inception angles, simulated 1 millisecond apart. 
By varying the four fault parameters under different loading conditions of the system,
25,560
fault cases are simulated, all triggering $LCDR_{AC}$ to trip. In each case, the LCDR measurements are recorded, similar to FDIAs. 
Both faults and FDIA cases are properly labeled, shuffled, and combined into one dataset.
Afterward, 20\% of this dataset is held out for testing, and the rest is used for the training phase. 
For ease, the training and testing datasets are denoted $\mathbf{D}_{train}$ and $\mathbf{D}_{test}$, respectively.
This splitting is done while ensuring 
that the ratio of fault to FDIA cases is the same in both
$\mathbf{D}_{train}$ and $\mathbf{D}_{test}$.
\subsection{Model Training and Testing Procedure}\label{section:Process}
Four models, namely MLP, CNN, ResNet, and  LSTM, are used to secure  $LCDR_{AC}$ against FDIAs following only the approach proposed in previous literature, e.g., \cite{mypaper_TII}, as discussed earlier. A summary of the models' architectures
and 
details 
can be found in \cite{tsai}.
Each of the four models is trained on the original training dataset $\mathbf{D}_{train}$, generated in Section \ref{section:Dataset},  to learn to distinguish between internal faults and general FDIAs on  $LCDR_{AC}$. (Note that these general FDIAs are performed without considering the existence of any FDIA detection system, i.e., remote measurements of $LCDR_{AC}$ are manipulated just to satisfy the LCDR's operating criterion.) 
Next, the four models are tested twice: one time on the original testing dataset $\mathbf{D}_{test}$, and the other time on $\mathbf{D}_{test}'$ which is the testing dataset that contains adversarial FDIAs generated using Algorithm \ref{alg:adversarial_attack_framework}.
In this experiment, an $\epsilon$ value of 0.5 is used along with 5 FGSM iterations.
\subsection{Evaluation Metrics}
The role of the DLS is to detect any FDIA—whether original or adversarial—that targets the augmented LCDR while ensuring the LCDR's sensitivity to faults remains minimally impacted.
To thoroughly evaluate the performance of the FDIA detection models and to account for any imbalance between the number of fault samples and FDIA samples, the following standard statistical metrics are used. These are the Accuracy, Precision, Recall, and F1-score  \cite{lecun2015deep}.
Moreover, to evaluate the robustness of the FDIA-detection models under adversarial attacks,
the Fooling Rate ($FR$) metric is used, defined as
\begin{equation}
\begin{aligned}
FR\text{ }(\%) = 
\textcolor{white}{..............................................................}   \\ 100 \cdot \frac{\sum_{i=1}^{N_{FDIAs}} \mathbb{I}(y_{i} \neq \hat{y}(\mathbf{x}') \ \text{and} \ \hat{y}(\mathbf{x}') = y_{\text{target}})}{N_{FDIAs}}
\end{aligned}
\end{equation}
\noindent where $N_{FDIAs} $ is the total number of FDIAs in the given dataset, and $\mathbb{I}$ denotes the indicator function, whose value is 1 if the enclosed condition is true and 0 otherwise. The  $FR$ measures the percentage of adversarial FDIAs that were misclassified by the model to the desired target class of faults and also triggered the LCDR to trip. Therefore, this metric reflects the effectiveness of the adversarial attacks by quantifying how often the adversarial perturbations caused a misclassification by the FDIA detection model. 
\subsection{Results and  Discussions}\label{section:Main_Results}
\subsubsection{Results Before Considering Adversarial Attacks}\label{section:Acc_before_adv} 
\begin{figure*}[t!]
\centering
\includegraphics
[width=2.0\columnwidth] {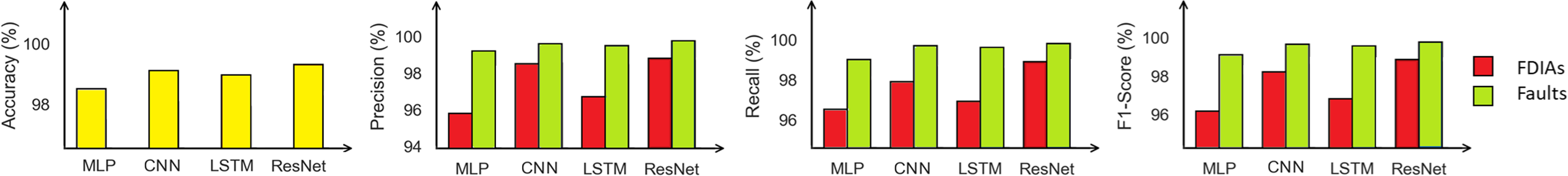}
\caption{ Original performance metrics \textit{before} adversarial attacks}
  \label{fig:Original_Results}
\end{figure*}

\begin{figure}[t!]
\centering
\includegraphics
[width=1.0\columnwidth] {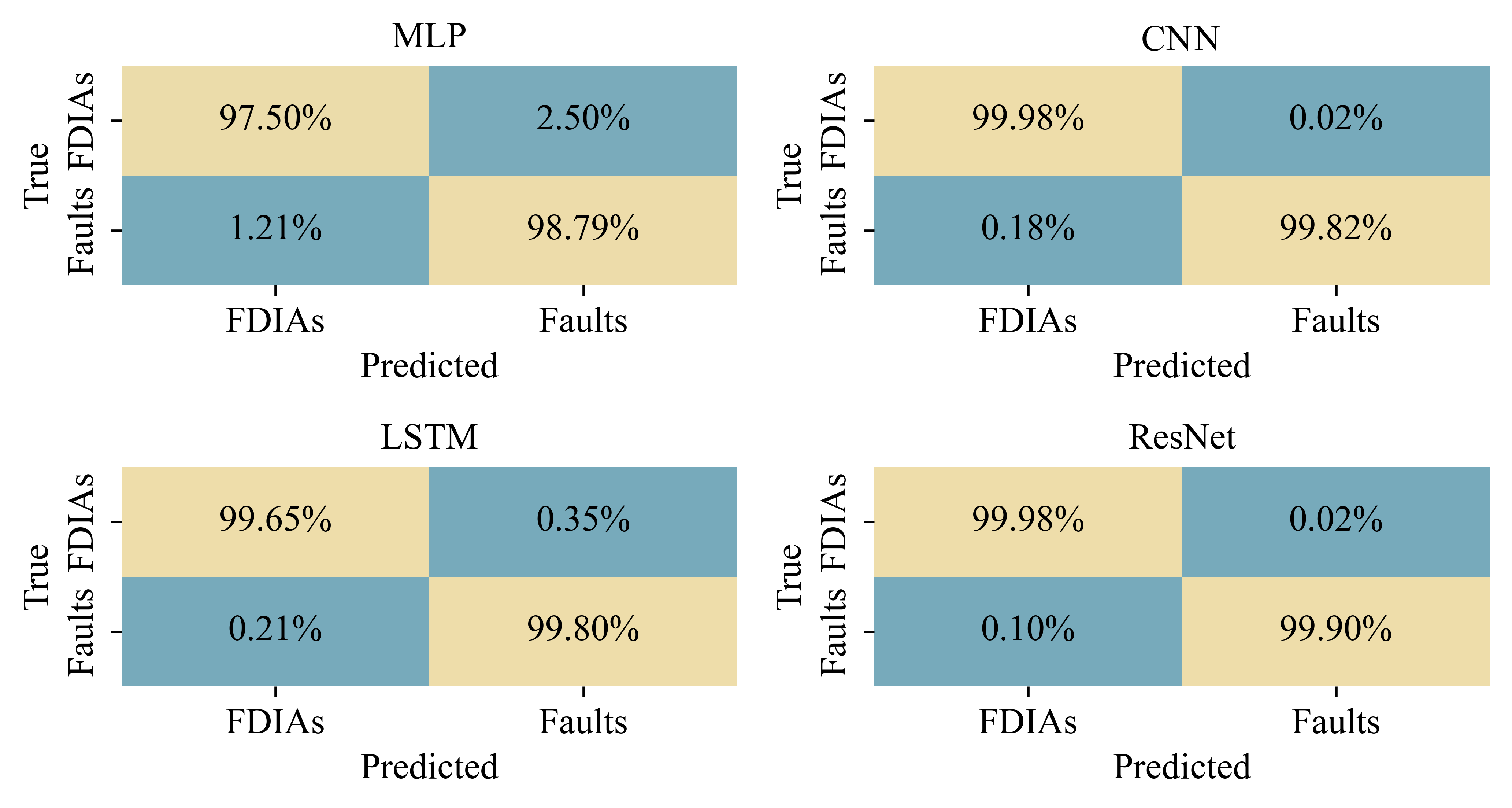}
\caption{Confusion matrices without adversarial attacks} 
  \label{fig:confusion_matrices_Original}
\end{figure}

Figures
\ref{fig:Original_Results} and \ref{fig:confusion_matrices_Original}
depict the performance results of each of the DLS models if implemented in $LCDR_{AC}$ when tested on $\mathbf{D}_{test}$.
It can be noticed that all the models can accurately distinguish between faults and FDIAs originally. The tables also show that the ResNet model is slightly more accurate than all the other models, followed by CNN and then LSTM. MLP, which is the technique proposed by recent related works, e.g., \cite{mypaper_TII}, is outperformed by all other models. 
%
%
%
\subsubsection{Results Under Adversarial Attacks}\label{section:Acc_after_adv}
\begin{figure*}[t!]
\centering
\includegraphics
[width=2.0\columnwidth] {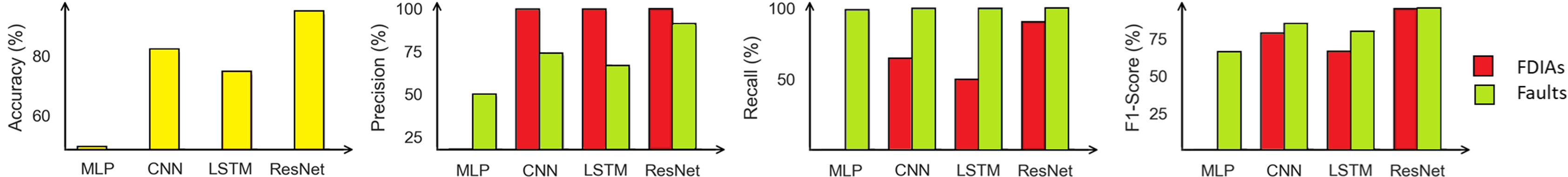}
\caption{ Performance metrics under adversarial attacks}
  \label{fig:Adversarial_Results}
\end{figure*}

\begin{figure}[t!]
\centering
\includegraphics
[width=1.0\columnwidth] {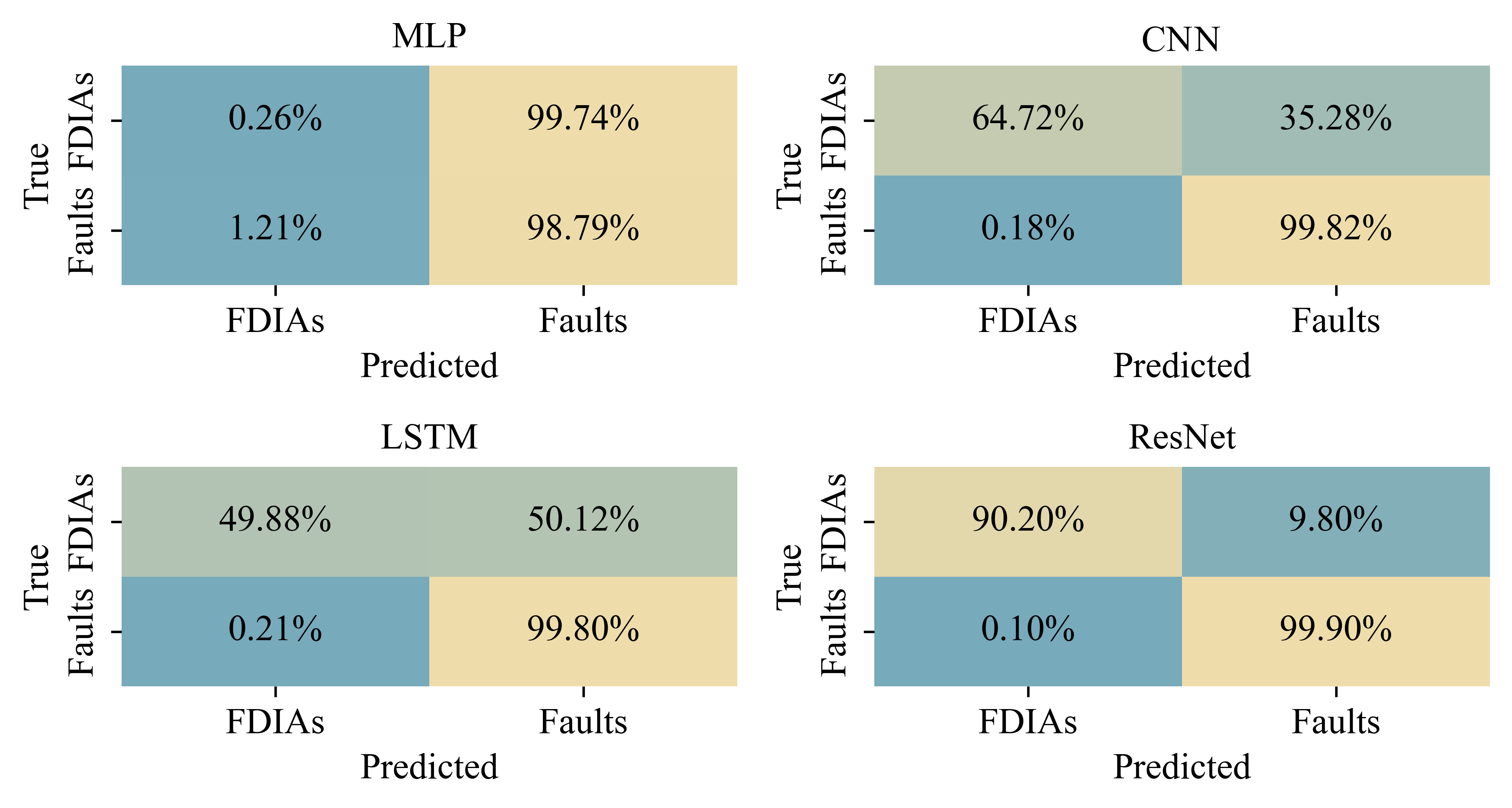}
\caption{Confusion matrices 
 under adversarial attacks
 }
\label{fig:confusion_matrices_Under_Attack}
\end{figure}

Under adversarial conditions, Figures \ref{fig:confusion_matrices_Under_Attack} and \ref{fig:Adversarial_Results} demonstrate that all \Ablue{investigated} DLS models exhibit varying degrees of vulnerability, as evidenced by the increased percentage of undetected FDIAs that trigger $LCDR_{AC}$. The results reveal notable differences in robustness across models.
\Ablue{The MLP model is the most susceptible to adversarial attacks, with 99.74\% attack success rate. In other words,
 99.74\% of adversarial FDIA samples are misclassified as faults.}
This reflects the shallow architecture's inability to handle adversarial perturbations, making it particularly vulnerable.
In contrast, ResNet shows the highest resilience, successfully detecting 90.2\% of adversarial FDIA samples. Its deep architecture and use of residual connections likely contribute to this robustness, enabling better generalization under adversarial conditions. CNN and LSTM perform moderately, with 64.72\% and 49.88\% of adversarial FDIA samples correctly detected, respectively. However, both models still show considerable vulnerability, especially as perturbations increase.
Table \ref{tab:poisoned_samples} highlights the FGSM's success rate in generating adversarial samples that bypass each model. The MLP model was the most compromised, with 97.24\% of FDIA samples successfully manipulated, while ResNet maintained the strongest defense, with only 9.78\% of samples bypassing it. CNN and LSTM fall between these two extremes, with poisoning rates of 35.26\% and 49.77\%, respectively.
Evidently, adversarial attacks severely degrade the FDIA detection performance of all models.
The high variability across model architectures highlights the critical need to account for adversarial robustness when designing FDIA detection systems for power grids. While MLP is particularly vulnerable, more complex models like ResNet offer greater resilience, albeit still not impervious to sophisticated attacks. These findings stress the importance of enhancing the robustness of these models to adversarial attacks, as discussed in the following section.
%
%
\begin{table}[!t]
\centering
\begingroup
\caption{Percentages of successful adversarial samples}
\label{tab:poisoned_samples}
\begin{tabular}{c | c  c c c}
\Xhline{3\arrayrulewidth}
\rule{0pt}{3ex}  \textbf{Model} 
&MLP 
&CNN  
&LSTM
&ResNet
\\
\rule{0pt}{3ex}  \textbf{ Poisoned Samples} 
& 97.24\% %
& 35.26\% %
& 49.77\%    
& 9.78\%  
\\
\Xhline{3\arrayrulewidth}
\end{tabular}
\endgroup
\end{table}
%
\section{Sensitivity Analyses}\label{section:Sensitivity}
\subsection{Impact of Varying the Magnitude of the Perturbation}\label{section:Sensitivity_epsilon}
\begin{figure}[t!]
\centering
\includegraphics
[width=1\columnwidth] {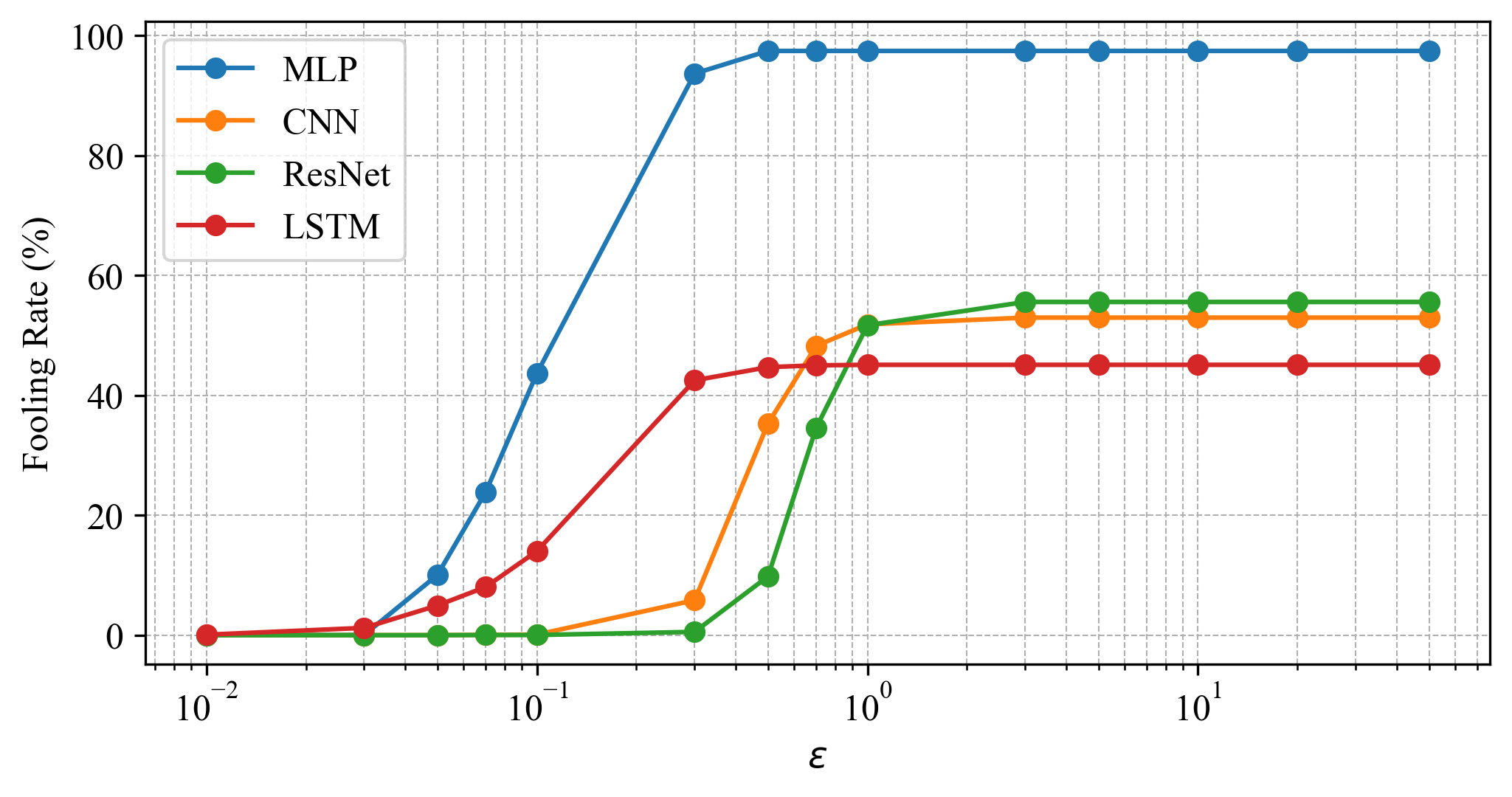}
\caption{ $FR$ at different $\epsilon$ values.}
  \label{fig:Sensitivity_epsilon}
\end{figure} 
This subsection evaluates the impact of the FGSM FDIA on the various DL models by varying the $\epsilon$ parameter, which controls the magnitude of input perturbations. The $FR$ is measured for the cases when $LCDR_{AC}$ is augmented by MLP, CNN, ResNet, and LSTM models. Fig. \ref{fig:Sensitivity_epsilon} illustrates the relationship between $\epsilon$ and $FR$ for each model. The results reveal a hierarchy in model robustness against FGSM-based FDIAs.
As shown in the figure, higher $\epsilon$ values lead to a significant increase in FR for all models, with MLP being the most vulnerable. Even at moderate perturbations ($\epsilon$ = 0.3), MLP’s FR rapidly rises to 93.66\%, eventually reaching 97.44\% at $\epsilon$ $\ge$ 0.5.
In contrast, CNN, ResNet, and LSTM exhibit greater resilience, with FRs below 55. 5\% for larger values $\epsilon$, demonstrating their ability to withstand adversarial attacks better than MLP. However, even these models are not immune to stronger perturbations.
%
%
\subsection{Impact of Varying the 
Number of Attack Iterations}
This subsection evaluates the impact of increasing the maximum number of allowable FGSM iterations. Herein, the maximum iterations are increased 10-fold, i.e., to become 50. A new testing dataset with adversarial FDIA samples is generated, following to the procedure explained in the previous sections.  
Our results show that the $Fooling\text{ } rate$ increased by 2.56\% in the case of the MLP model. Conversely, there is no observed change in the $Fooling\text{ } rate$ for the cases of the rest of the models. These results reflect the significant impact adversarial attacks can have on DL-based FDIA detectors in LCDRs, even after a single iteration.
%
%
%
%
\section{ Adversarial Training for Proactively Defending LCDRs Against Adversarial Attacks}\label{section:Defense}
This section employs adversarial training, a proactive and intuitive defense strategy, to improve the resilience of the DLS models to adversarial attacks \cite{goodfellow2014explaining,tian2021adversarial}. In this approach, the DLS models are updated by exposing them to a wide number of adversarial samples, so they can correctly classify them as FDIAs in the future. The proposed adversarial training process, depicted in  \ref{alg:adversarial_training}, involves two stages for each model. Firstly, Algorithm \ref{alg:adversarial_attack_framework} is applied to the original training dataset $\mathbf{D}_{train}$ to produce a set of adversarial FDIA samples. These new samples are then used to augment $\mathbf{D}_{train}$, resulting in a training dataset with adversarial samples, denoted $\mathbf{D}_{train}'$.
Next, the model's weights are updated by training it on the new $\mathbf{D}_{train}'$, resulting in a model robust to adversarial FDIAs, denoted $\mathbf{f}_{\bm{\theta} r}$.
\begin{algorithm}[t!]
\caption{Proposed adversarial training defense}
\label{alg:adversarial_training}
\KwIn{$\mathbf{f}_{\bm{\theta}}$: DLS model, $\mathbf{D}_{train}$: original training dataset, $N_{\text{ep}}$: no. of adversarial training epochs}
\KwOut{$\mathbf{f}_{\bm{\theta} r}$: A robust FDIA detection model}
\SetNlSty{textbf}{}{} %
\textbf{Initialize:}  $\mathbf{D}_{train}'$ $\gets$ $\mathbf{D}_{train}$, $\mathbf{f}_{\bm{\theta} r}$ $\gets$ $\mathbf{f}_{\bm{\theta}}$\;%
\textbf{Generate Adversarial Samples:}\;
\For{$i \gets 1$ \textbf{to} (number of samples in $\mathbf{D}_{train}$)}{
    Extract the $i$th sample $\mathbf{x}$ and its label  $y$;\\
    \If{ ($\hat y (\mathbf{x}) =  1$) \text{\textbf{and}} ($y =  1$)}{
        Attempt to generate an adversarial sample out of $\mathbf{x}$ using Algorithm \ref{alg:adversarial_attack_framework}\;
        \If{ Algorithm \ref{alg:adversarial_attack_framework} returns a \textbf{successful} adversarial FDIA sample $\mathbf{x}'$}{
            \textbf{Label $\mathbf{x}'$ as FDIA:} $y' = 1$, and \\ \textbf{Augment training dataset:} Add $\mathbf{x}'$ and $y'$ to $\mathbf{D}_{train}'$\;
        }
    }
}
\textbf{Re-training Process:}\;
\For{$epoch \gets 1$ \textbf{to} $N_{\text{ep}}$}{
    Train $\mathbf{f}_{\bm{\theta} r}$ on  $\mathbf{D}_{train}'$\;
}
\Return{$\mathbf{f}_{\bm{\theta} r}$ \tcp*[l]{A robust DLS model}}
\end{algorithm}
\subsection{Adversarial Training Results}
\begin{figure}[t!]
\centering
\includegraphics
[width=1.0\columnwidth] {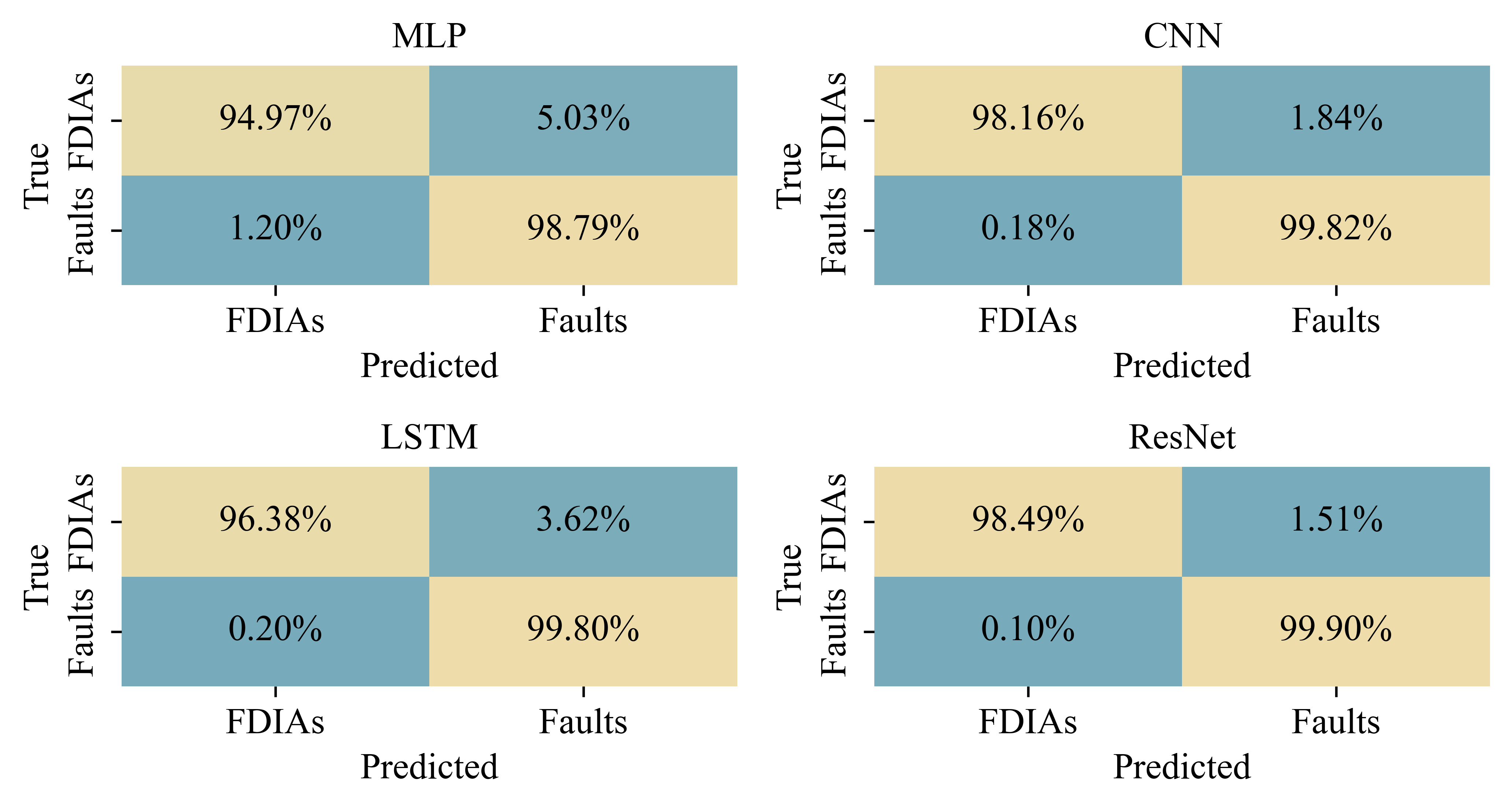}
\caption{Confusion matrices 
after adversarial training}
\label{fig:confusion_matrices_After_Training}
\end{figure}

The aforementioned methodology is applied to obtain four robust DLS models. For each model, adversarial training is performed for 10 additional epochs using the same $\epsilon$ values used in Section \ref{section:Sensitivity}. 
Each of the four new models is then tested against an adversarial testing Dataset $\mathbf{D}_{test}'$ generated using 
the approach described earlier in Section \ref{section:Simulation_Results}.
Fig. \ref{fig:confusion_matrices_After_Training} shows the confusion matrices after adversarial training. Comparing these results with those depicted earlier in
Fig. \ref{fig:confusion_matrices_Under_Attack} reveals that adversarial training significantly improves the robustness of the DLS models to adversarial attacks, indicated by the enhanced FDIA detection accuracy. 
%
%
\subsection{Inference Times and  Real-Time Simulation}\label{section:Time}
To ensure their adequacy, this subsection also measures the inference time of the robust DLS models trained in the previous subsection. This is performed on a PC with an Intel® Xeon® E5-2683 v4 processor running at 2.1 GHz and with 125 GB of RAM.
To measure the prediction time, each model is tested on 1000 cases of different faults and FDIAs, and then the average is calculated.
Table \ref{tab:prediction_times} shows that the time taken by the DLS to detect FDIAs.
\Ablue{Our results show that after the LCDR is triggered to trip, the DLS takes an additional duration of 0.755 to 1.37 milliseconds to detect FDIAs (or confirm faults),  depending on the model used. Consequently, the total detection and confirmation time becomes 25–33.33 ms (1.5 to 2 power cycles \cite{GELCDR}) for the LCDR, plus 0.755–1.37 ms for the DL model. This slight increase is minimal and, therefore, considered acceptable.}

\begin{figure}[t!]
\centering
\includegraphics 
[width=1\columnwidth] {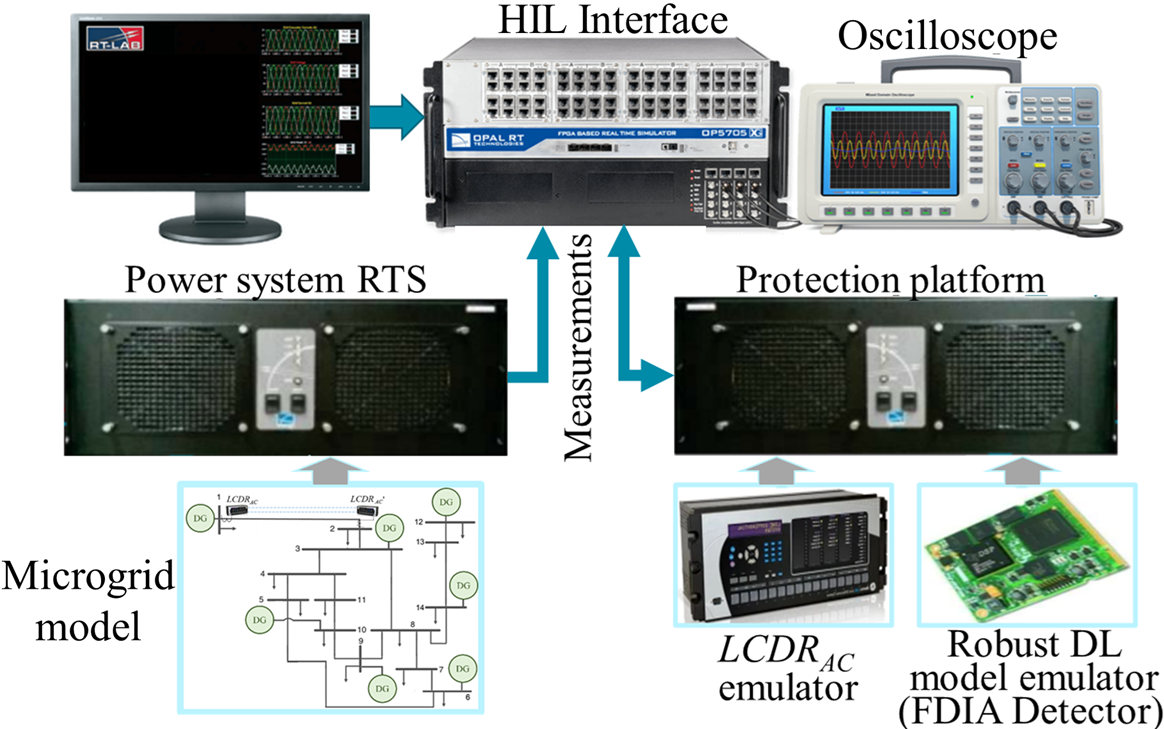}
\caption{Real-time simulation testbed}  
  \label{fig:RTS}
\end{figure}

\begin{figure}[t!]
\centering
\includegraphics [width=0.75\columnwidth] {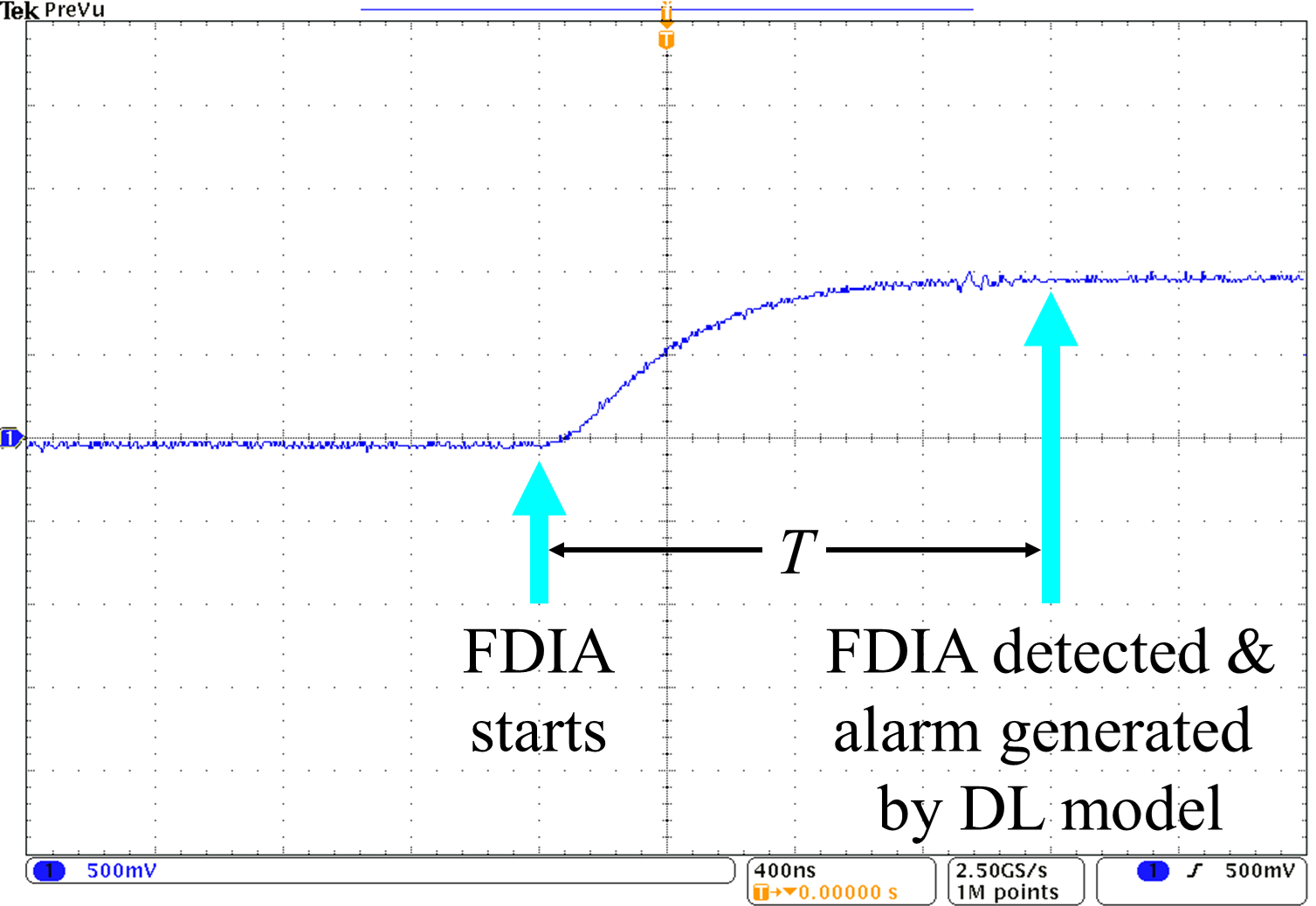}
\caption{  Oscilloscope snapshot, FDIA detection time.}  
  \label{fig:Adv_OPAL_Oscilloscope}
\end{figure}

For further verification, we implement a robust model in the real-time testbed. 
illustrated in Fig. \ref{fig:RTS}. The setup consists of:  1) a Real-Time Simulator (\Ablue{RTS}) with a Hardware-in-the-Loop (HIL) configuration,  2) an
oscilloscope, and 3) a PC. The \Ablue{RTS} system, based on the OP5700 RCP/HIL, leverages a group of high-end field-programmable gate arrays and Intel Xeon E5 quad-core processors running at speed of
3 GHz \cite{OP5700}.
\Ablue{As illustrated in  Fig.~\ref{fig:RTS}, in our RTS setup, a power system simulator and a protection platform run on separate CPU cores connected in a loop. 
In this experiment, a similar four-cycle observation window and three-phase local and remote current are measured, sampled at 1 kHz. 
The FDIA-detection emulator running at 2.3 GHz is utilized.
The \Ablue{FDIA} detection signal is sent from the \Ablue{RTS} to the oscilloscope via one of the input/output ports for monitoring. 
To enhance visibility, the time scale of the oscilloscope’s display is configured at 400ns.}
\Ablue{In this experiment, we focus on $LCDR_{AC}$ similar to the previous sections. This LCDR is augmented with an MLP robust DL attack detection model. The protection platform, which emulates $LCDR_{AC}$ augmented with the detection model, receives time-stamped local measurements directly from the power system simulator through the HIL interface. Similarly, it obtains time-stamped  sampled value (SV) remote measurements from the power system simulator, emulating the scenarios where the LCDR receives SV measurement packets from the remote LCDR.}
\Ablue{Using this setup, an adversarial FDIA case is performed. 
To simulate this FDIA, the remote measurements are deliberately manipulated upon the FDIA onset within the power system simulator before the measurements are sent to the HIL interface, thereby replicating a scenario in which a cyber-attacker disrupts the flow of SV packets and injects falsified payloads into those packets reaching $LCDR_{AC}$. This configuration enables us to validate the MLP model by measuring the time required by the robust DL FDIA-detection model to detect an attack (or fault) in real time; our results confirm that the FDIA is detected and an alarm is generated in under 2 milliseconds,  as shown in Fig. \ref{fig:Adv_OPAL_Oscilloscope}, demonstrating acceptable real-time performance and preserving the LCDR’s speed merit while enhancing its resilience against cyberattacks.}

\Ablue{Utilizing a slower processor shall increase the FDIA detection time. 
Further,
while our test setup accurately emulates the measurement acquisition of modern LCDRs, a practical deployment in an operational power grid would involve additional complexities, including
factors like protocol-specific latency, packet loss, or synchronization errors, which were not modeled in this paper as the focus is on minimizing the risk of LCDR false tripping under adversarial FDIAs. 
For example, factors such as synchronization errors between local and remote measurements of an LCDR can cause false operation of this LCDR, similar to the  goal of the FDIAs in this paper.
The proposed method compliments existing LCDR mechanisms such as disturbance detection that  delay the LCDR's tripping for a short period of time, e.g., one cycle, to avoid instantaneous maloperation of LCDRs due to external disturbances \cite{zimmerman2013practical}. }

\begin{table}[!t]
\centering
\begingroup
\caption{Attack detection times }
\label{tab:prediction_times}
\begin{tabular}{c | c c c c}
\Xhline{3\arrayrulewidth}
\rule{0pt}{3ex} \textbf{Model} & MLP& CNN &  LSTM & ResNet 
\\ 
\rule{0pt}{3ex} \textbf{Prediction Time (ms)} & 0.755 & 0.769 & 1.37 & 1.04 
\\
\Xhline{3\arrayrulewidth}
\end{tabular}
\endgroup
\end{table}
\section{Conclusion}\label{section:Conclusion}
 This paper unveils the vulnerability of DL-based FDIA detection systems in LCDRs to adversarial attacks. 
It presents an adversarial attack framework that employs the FGSM to exploit vulnerabilities in the DLS by carefully perturbing the targeted LCDR's remote measurements. This manipulation causes the DLS to misclassify an adversarial FDIA sample as a legitimate fault, triggering the LCDR to trip. The ability of adversarial samples to deceive both detection models and the LCDR underscores the potential for inducing cascading failures in the power system.
A comprehensive evaluation of multiple DL models, including MLP, CNN, LSTM, and ResNet, was performed. \Ablue{Our results confirm that while all the investigated models initially perform well, their performance degrades significantly when subjected to adversarial FDIAs, with attack success rates ranging from 9.78\% to over 99\% across different architectures and varying magnitudes of perturbation.} Of particular concern is the high vulnerability of the MLP model, which has been proposed in several related works for LCDR cyberattack detection, while other, more complex models exhibited a lower degree of vulnerability. %
Furthermore, the paper proposed adversarial training as a proactive defense mechanism. \Ablue{Our results confirm that the proposed adversarial training approach not only significantly improves the models' ability to detect adversarial FDIAs but also maintains high detection accuracy for legitimate faults. }
Therefore, we recommend that utilities planning to incorporate DL models, particularly for FDIA detection, ensure the robustness of these models against adversarial attacks by utilizing the approaches proposed in this paper.
\Ablue{Future work includes evaluating the proposed scheme under various types of synchronization errors, communication delays and  protocol-specific vulnerabilities.}

\bibliographystyle{IEEEtran} 
\bibliography{TII-Articles-LaTeX-template/Manuscript}

\begin{IEEEbiography} [{\includegraphics[width=1in,height=1.25in,clip,keepaspectratio]{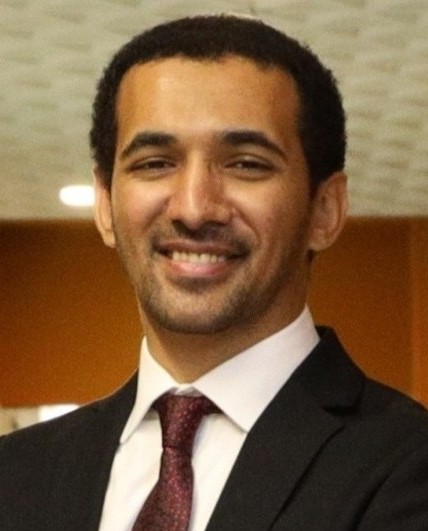}}]{Ahmad Mohammad Saber}
(M’21) received the B.Sc. degree from Ain Shams University, Egypt, in 2016, the M.Sc. degree from Cairo University, Egypt, in 2019, and the Ph.D degree from Khalifa University, UAE, in 2024. 
He is currently a Postdoctoral fellow at the University of Toronto, ON, Canada. 
Ahmad's research received the UAE Excellence and Creative Engineering Award in 2024.
Since 2016, he has held technical and commercial roles in several industries including power transformers manufacturing, water and wastewater treatment, and access control and security systems. In 2023, he was an international visiting graduate student at the University of Toronto, ON, Canada.  His current research interests include cyber-physical security, AI applications and security, power system protection, distributed generation, and renewable power planning and integration. 
Dr. Saber is a reviewer for multiple IEEE Transactions journals.

\end{IEEEbiography}

\begin{IEEEbiography} [{\includegraphics[width=1in,height=1.25in,clip,keepaspectratio]{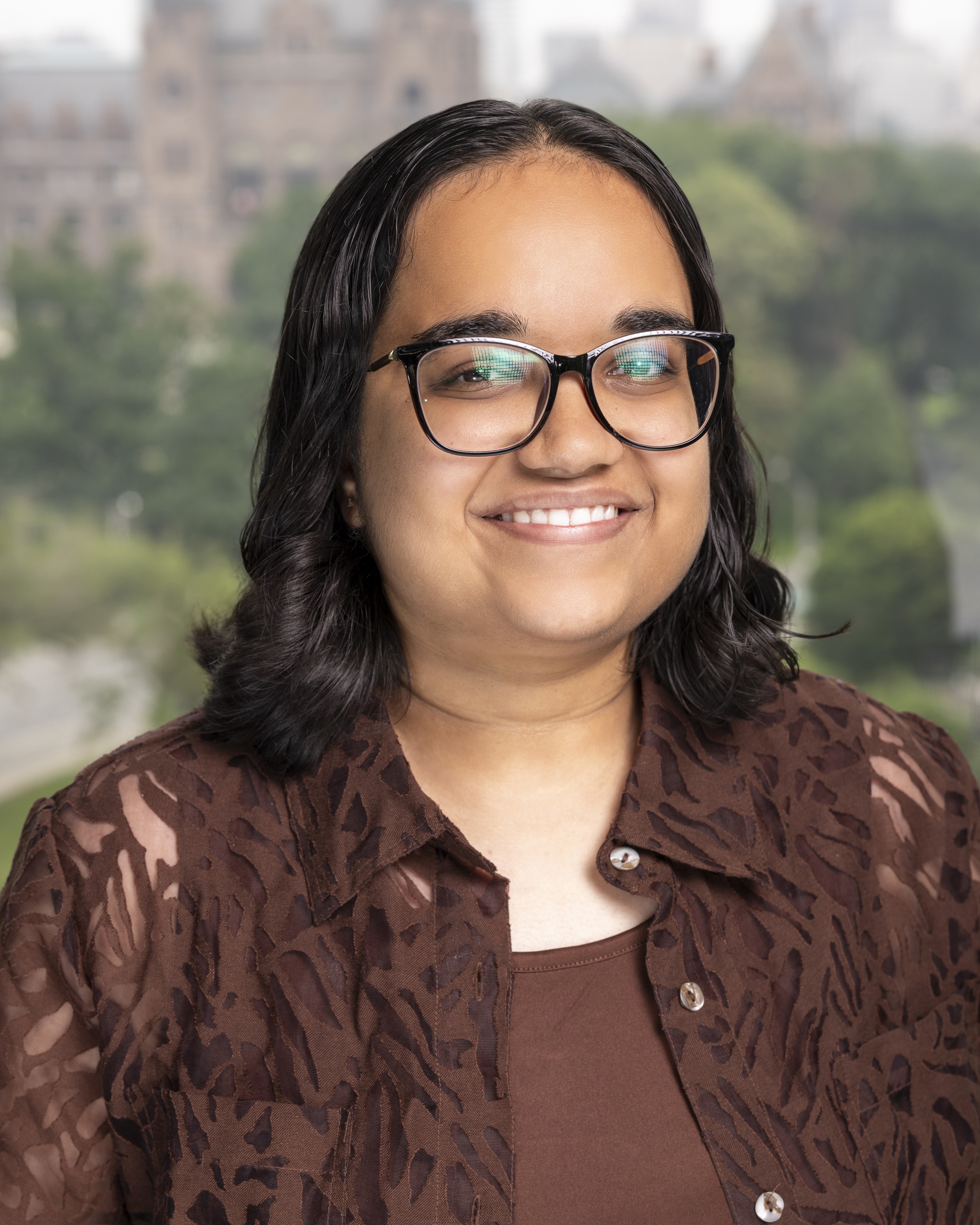}}]{Aditi Maheshwari} is pursuing a M.A.Sc. in Computer Engineering from the University of Toronto (UofT), ON, Canada, focused on classical and quantum machine learning for power grid cybersecurity. She holds a B.A.Sc. in Engineering Science (Robotics major, AI minor) also from the University of Toronto. Through research, industry internships, and teaching roles, she has engineering experience across multiple domains—including healthcare, climate action, security, software, business analytics, and consumer products. She has been recognized as the winner of the \textit{Rising Star in AI - Young Role Model of the Year} Award by Women in AI North America. Her research interests lie in leveraging emerging technology for interdisciplinary challenges.

\end{IEEEbiography}

\begin{IEEEbiography} [{\includegraphics[width=1in,height=1.25in,clip,keepaspectratio]{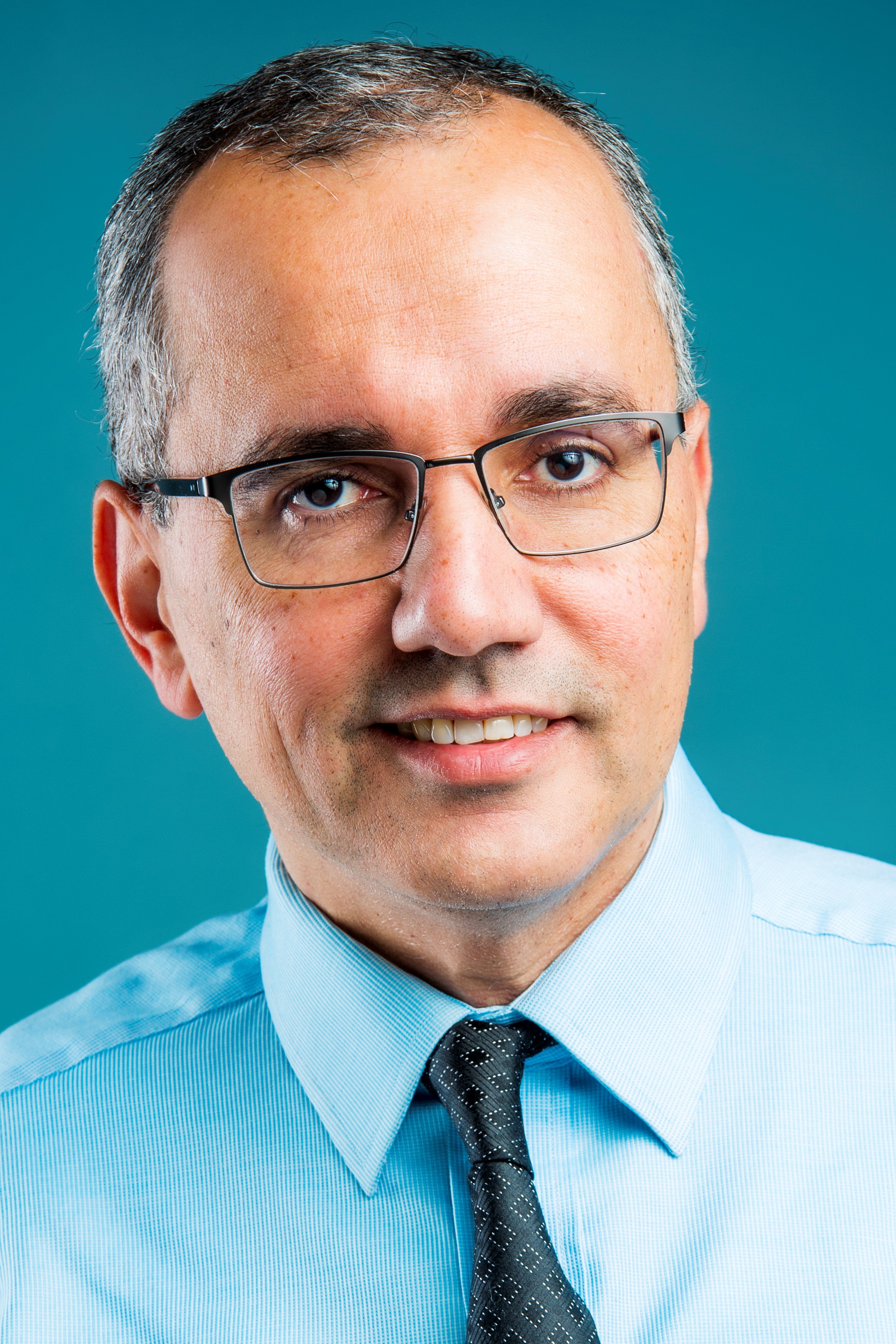}}]{Amr Youssef}
(SM’06) received the B.Sc. and M.Sc. degrees from Cairo University, Cairo, Egypt, in 1990 and 1993, respectively, 
and the Ph.D. degree from Queens University, Kingston, ON, Canada, in 1997. He was with Nortel Networks, the Center
for Applied Cryptographic Research, University of
Waterloo, IBM, and also with Cairo University. He is
currently a Professor with the Concordia Institute for
Information Systems Engineering, Concordia University, Montreal, Canada. 
He has authored over 300 referred journal and conference publications in areas related to his research interests. 
His current research interests include information security, and cyber-physical systems security. 
Dr. Youssef  served on over 100 technical program committees of cryptography and data security conferences. 
He was the co/chair for Africacrypt 2010, Africacrypt 2020, the conference Selected Areas in Cryptography (SAC 2014, SAC 2006, and SAC 2001).
\end{IEEEbiography}

\begin{IEEEbiography} [{\includegraphics[width=1in,height=1.25in,clip,keepaspectratio]{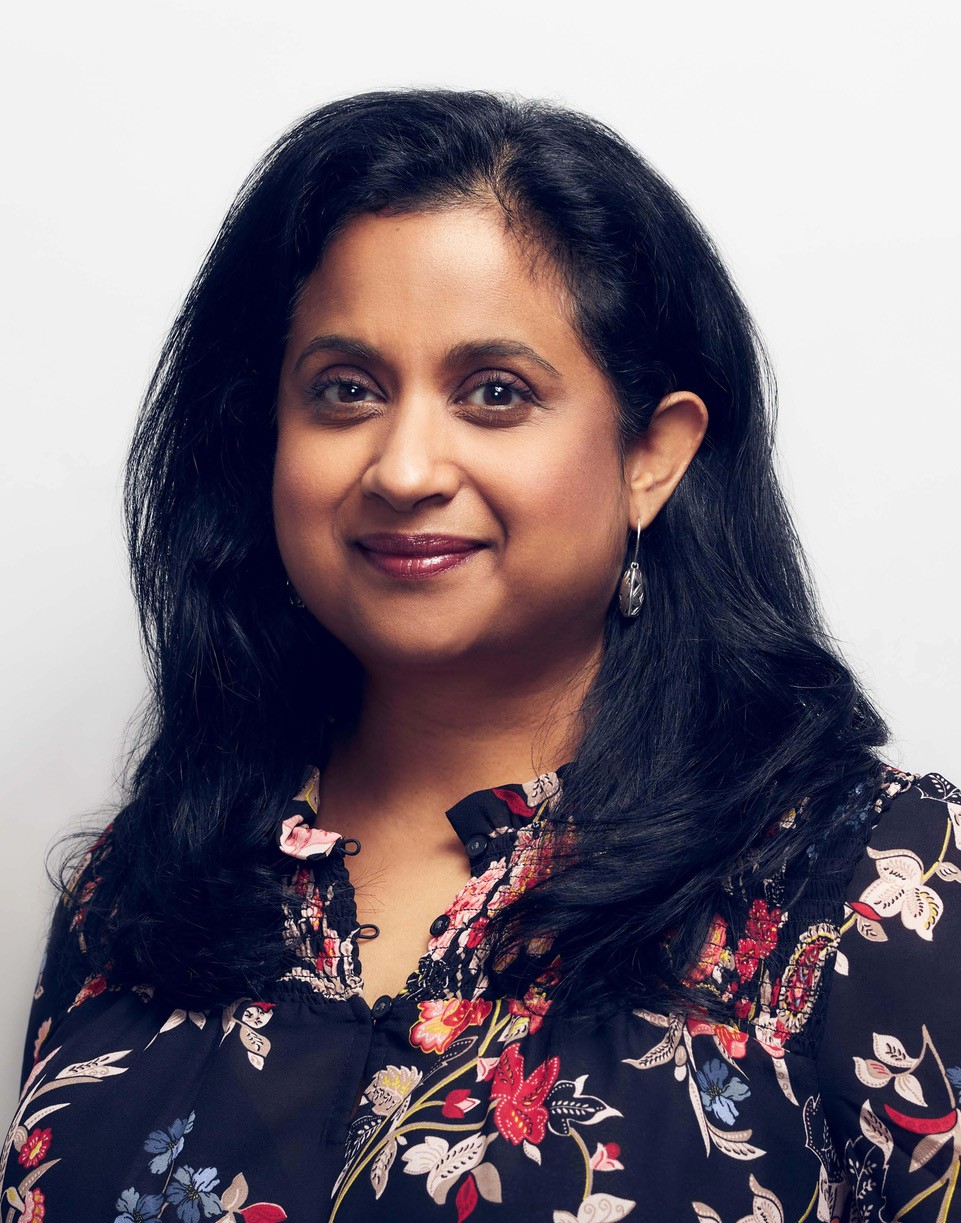}}]{Deepa Kundur}
(F’15) is the Canada Research Chair in Cybersecurity of Intelligent Critical Infrastructure and Chair of The Edward S. Rogers Sr. Department of Electrical \& Computer Engineering at the University of Toronto. A Toronto native, she received her B.A.Sc., M.A.Sc., and Ph.D. degrees—all in Electrical and Computer Engineering—from the University of Toronto in 1993, 1995, and 1999, respectively.

Professor Kundur’s research interests include cybersecurity of smart grid systems, autonomous electric vehicles, and psychiatric informatics. She has authored over 200 journal and conference publications and is regarded as a recognized authority on cybersecurity issues.

She has held several prominent leadership roles in the research community, including Honorary Chair of the 2021 IEEE Electric Power and Energy Conference, TPC Co-Chair of the 2023 IEEE SmartGridComm, and Publicity Chair for ICASSP 2021. She has also served in various executive organization capacities for flagship events such as IEEE GlobalSIP, IEEE ICC, ACM e-Energy, and IEEE GLOBECOM, particularly in tracks and symposia focused on smart energy systems, resilient infrastructures, and cyber-physical security.

Professor Kundur is a Fellow of the IEEE, the Canadian Academy of Engineering, and the Engineering Institute of Canada. She is also a Senior Fellow of Massey College.
\end{IEEEbiography}

\end{document}